\pdfoutput=1 
\documentclass[journal]{IEEEtran}

\usepackage{cite}
\usepackage{amsmath,amssymb}
\usepackage{algorithm}
\usepackage{algpseudocode} 
\usepackage{graphicx}
\usepackage{textcomp}
\usepackage{xcolor}
\usepackage{colortbl} 
\usepackage{multirow} 
\usepackage{pifont} 
\usepackage{cite}
\usepackage{algpseudocode} 
\usepackage{booktabs} 
\usepackage{sidecap}
\usepackage{subcaption}
\usepackage{tikz}
\usepackage{soul}

\usepackage{graphicx}

\usetikzlibrary{fadings,shadings,shapes}

\definecolor{deemph}{gray}{0.6}

\definecolor{customgreen}{RGB}{0, 153, 0} 

\definecolor{lossred}{RGB}{255, 0, 0}

\newcommand{\cmark}{\ding{51}}%
\newcommand{\xmark}{\ding{55}}

\newcommand{\yes}{\textcolor{darkgray}{\ding{51}}}
\newcommand{\no}{\textcolor{lightgray}{\ding{55}}}

\newcommand{\numbersBlue}[1]{\textcolor{blue}{\textbf{#1}}}

\definecolor{mygray}{gray}{.92}
\definecolor{customviolet}{RGB}{148, 0, 211}
\definecolor{cvprblue}{rgb}{0.21,0.49,0.74}
\definecolor{Green}{rgb}{0.85882353, 0.90980392, 0.84705882}
\newcommand{\etal}{\textit{et al.}}

\makeatletter
\newcommand{\thickhline}{%
    \noalign {\ifnum 0=`}\fi \hrule height 1pt
    \futurelet \reserved@a \@xhline
}
\makeatother

\usepackage{hyperref}
\hypersetup{
    colorlinks=true,
    linkcolor=black,
    citecolor=cvprblue,
    filecolor=black,
    urlcolor=cvprblue,
}

\title{GenMix:  Effective Data Augmentation with Generative Diffusion Model Image Editing}

\author{
    Khawar Islam, Muhammad Zaigham Zaheer, Arif Mahmood, Karthik Nandakumar, and Naveed Akhtar%
    \thanks{Khawar Islam and Naveed Akhtar are with the School of Computing and Information Systems, The University of Melbourne, Australia.}
    \thanks{Muhammad Zaigham Zaheer and Karthik Nandakumar are with Mohamed bin Zayed University of Artificial Intelligence, UAE.}
    \thanks{Arif Mahmood is with Information Technology University, Punjab, Pakistan.}
    \thanks{Corresponding to: Khawar Islam (Email: khawari@student.unimelb.edu.au)}
}

\begin{document}

\IEEEtitleabstractindextext{
\begin{abstract}
Data augmentation is widely used to enhance generalization in visual classification tasks. However, traditional methods struggle when source and target domains differ, as in domain adaptation, due to their inability to address domain gaps. This paper introduces GenMix, a generalizable prompt-guided generative data augmentation approach that enhances both in-domain and cross-domain image classification. Our technique leverages  image editing to generate augmented images based on custom conditional prompts, designed specifically for each problem type. By blending portions of the input image with its edited generative counterpart and incorporating fractal patterns, our approach mitigates unrealistic images and label ambiguity, improving performance and adversarial robustness of the resulting models. Efficacy of our method is established with extensive experiments on \textcolor{black}{eight} 
public datasets for general and fine-grained classification, in both in-domain and cross-domain settings.
Additionally, we demonstrate performance  improvements for self-supervised learning, learning with data scarcity, and adversarial robustness. As compared to the existing state-of-the-art methods, our technique achieves stronger performance across the board. 
\end{abstract}

}

\maketitle
\IEEEdisplaynontitleabstractindextext
\IEEEpeerreviewmaketitle

\newcommand{\our}{{GenMix}}

\section{Introduction}
\label{sec:intro}

Deep Neural Networks (DNNs) have achieved remarkable advancements across various computer vision tasks, such as image classification \cite{dosovitskiy2020image, he2016deep, liu2022swin, tan2019efficientnet}, object detection \cite{fang2021you, he2017mask, wang2021pyramid}, image captioning \cite{fei2022deecap, ke2019reflective}, human pose estimation \cite{liu2021deep, luo2021rethinking}, and image segmentation \cite{badrinarayanan2017segnet, strudel2021segmenter}. However, as the complexity of these tasks increases, so do the computational costs associated with modern DNN architectures, including larger memory footprints, model sizes, higher FLOPs, and longer inference times \cite{devries2017improved}. Recent models employ billions of parameters to enhance feature representation. This often leads to overfitting and challenges in generalization, especially when the training data is limited or poorly structured. These issues become even more pronounced in cross-domain settings \textcolor{blue}{\cite{wang2022resmooth, lin2023constrained}} , where models must adapt to divergent data distributions.

\par
\begin{figure*}[t]
    \centering
    \includegraphics[width=0.90\linewidth]{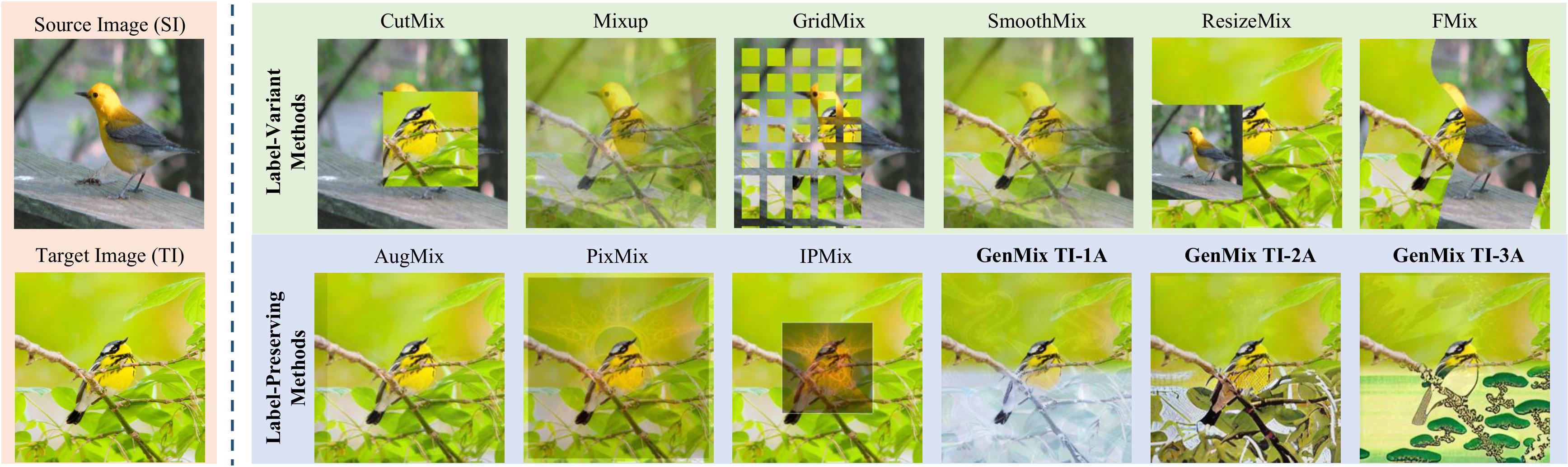}
    \vspace{-5pt}
    \caption{Representative examples of  augmentation techniques. \textbf{Top:} Conventional image mixing methods like CutMix \cite{cutmix}, Mixup \cite{mixup}, GridMix \cite{baek2021gridmix}, SmoothMix \cite{lee2020smoothmix}, ResizeMix \cite{qin2020resizemix} and FMix \cite{harris2020fmix} use interpolation or other strategies to mix source and target images, largely disregarding the natural appearance of the resulting image.
    \textbf{Bottom:} Our \our{} is a label-preserving technique like AugMix~{\cite{hendrycksaugmix}}, PixMix~{\cite{hendrycks2022pixmix}} and IPMix~{\cite{huang2023ipmix}}. However, it utilizes a generative diffusion model with custom tailored prompts for high quality augmentation. Three different outputs (1A, 2A and 3A) for the target are shown for GenMix.}  
    \label{fig:overview}
     \vspace{-10pt}
\end{figure*}

Image mixing based data augmentation techniques have proven  effective for enhancing the generalization of deep learning models \textcolor{blue}{\cite{cheung2023survey}}. A wide array of such techniques has been  introduced, including CutMix \cite{yun2019cutmix}, Mixup \cite{kim2020puzzle}, SnapMix \cite{huang2021snapmix}, Co-Mixup \cite{kim2020co}, SaliencyMix \cite{uddin2020saliencymix}, LGCOAMix \cite{dornaika2023lgcoamix}, GuidedMixup \cite{kang2023guidedmixup}, MixPro \cite{zhao2022mixpro}, TransMix \cite{chen2022transmix}, and PuzzleMix \cite{kim2020puzzle}. These methods typically involve combining portions of randomly selected images of corresponding labels through various mixing strategies to generate augmented samples - see top row in Fig.~\ref{fig:overview}. By employing linear interpolation or other blending techniques, they create novel training images to enrich the dataset for addressing potential model overfitting~\cite{cutmix, mixup, han2022yoco, qin2023adversarial, huang2021snapmix,islam2022recent}. 
\par
Though effective, image mixing based data augmentation encounters   challenges, such as the omission of salient regions and label ambiguities due to the random blending of image content \cite{kim2020puzzle}. To mitigate these issues, several approaches have incorporated saliency-driven methods, which prioritize the preservation of important regions by overlaying them onto less significant background areas of the target image \cite{kim2020puzzle, uddin2020saliencymix, kang2023guidedmixup, islam2022face}. While these techniques offer improvements, they introduce considerable computational overhead and rely on the accuracy of the saliency  methods, which is a limiting factor. Furthermore, by combining images from diverse categories, these techniques can still overlook crucial contextual information, leaving the problem of inadequate feature preservation unresolved.
\par
Diffusion Models (DMs) \cite{takagi2023high, du2023stable, luo2023camdiff, saharia2022palette, dhariwal2021diffusion, tao2024unsupervised} have recently emerged as powerful tools for image-to-image generation and editing. Some studies \cite{trabucco2023effective, azizi2023synthetic, moser2024diffusion} have also explored using DM generated images to augment training data. However, we empirically  observe limited performance gains from these approaches, with models trained on synthetic data occasionally underperforming compared to the baseline models without augmentation \cite{azizi2023synthetic}. This can be attributed to the limited control over  image generation of the DMs. Due to the model  sensitivity to conditional prompts, generating complex \emph{scenes}, \emph{layouts}, and \emph{shapes} becomes challenging \cite{zhang2023adding, islam2025context}. Poorly crafted prompts often yield unsuitable images for augmentation, which can lead to model adaption to an incorrect data distribution. This underscores the need for more precise prompt selection and better control mechanisms in the augmentation process to improve both domain generalization \cite{wang2024smooth} and cross-domain performance.

To address this, we propose GenMix, a \textit{gen}erlizable data augmentation technique that leverages \textit{gen}erative pre-trained diffusion models to compute diverse augmentation samples through carefully designed conditional prompts. 
\our{} creates hybrid images by blending original and generated content, maintaining essential semantics of the original image, see bottom row of  Fig.~\ref{fig:overview}. To further enrich structural diversity, we incorporate self-similarity fractals into the hybrid images, a strategy proven to enhance machine learning safety \cite{hendrycks2022pixmix,huang2024ipmix}. This blending mitigates overfitting and improves model performance. Our strategy is unlike any previous attempts of directly using DM images for augmentation~\cite{azizi2023synthetic, trabucco2023effective}. 
It effectively incorporates a range of operations in its components to achieve  outstanding eventual performance - see Table~\ref{table:compar_study}.  Our experiments show that \our{} surpasses current state-of-the-art (SOTA) augmentation methods by improving generalization and adversarial robustness. Additionally, it facilitates prompt-guided learning in both in-domain and cross-domain tasks, making it highly adaptable for diverse computer vision problems. Furthermore, our results confirm that \our{} is compatible with a wide range of datasets and it integrates seamlessly into existing SOTA frameworks.


The main contributions of this work are as follows: 


\begin{itemize}
    \item We introduce a novel data augmentation method, GenMix that leverages image-editing with generative diffusion models guided by curated conditional prompts. Our technique merges original and generated images, followed by blending self-similarity fractals, which reduces overfitting and ensures robust training.
    \item We explore the applicability of  \our{} to both in-domain classification and domain adaptation tasks, introducing sets of bespoke conditional prompt for enhanced adaptability. Notably, \our{} is the first method to address both in-domain and domain adaptation tasks effectively.
    \item We demonstrate with extensive experiments across eight datasets  that \our{} outperforms existing SOTA augmentation methods in general classification, fine-grained classification, adversarial robustness, and cross-domain tasks.
\end{itemize}

\begin{table*}[t]
\setlength{\tabcolsep}{2.3pt}
\centering
\caption{Comparison of data augmentation methods. \textit{Components} include central operations and inputs. \textit{Tasks} are the key problems for which the efficacy of the methods is established.}
\vspace{-5pt}






\label{table:compar_study}
\begin{tabular}{l|lcccccccccccccccc}
\toprule 
& Input   & \begin{tabular}[c]{@{}c@{}} CutMix \\ \end{tabular} 
& \begin{tabular}[c]{@{}c@{}} Mixup \\ \end{tabular} 
& \begin{tabular}[c]{@{}c@{}} GridMix \\ \end{tabular} 
& \begin{tabular}[c]{@{}c@{}} SMix \\  \end{tabular} 
& \begin{tabular}[c]{@{}c@{}} ResMix \\  \end{tabular} 
& \begin{tabular}[c]{@{}c@{}} FMix \\ \end{tabular} 
& \begin{tabular}[c]{@{}c@{}} AugMix \\ \end{tabular}  
& \begin{tabular}[c]{@{}c@{}} PixMix \\ \end{tabular} 
& \begin{tabular}[c]{@{}c@{}} IPMix \\  \end{tabular} 
& \begin{tabular}[c]{@{}c@{}} AutoMix \\ \end{tabular} 
& \begin{tabular}[c]{@{}c@{}} AdAutoMix \\ \end{tabular} 
& \begin{tabular}[c]{@{}c@{}} DiffuseMix \\ \end{tabular} 
& \begin{tabular}[c]{@{}c@{}} \textbf{Ours} \end{tabular}  \\
\midrule

\multirow{7}{*}{\rotatebox{90}{ \emph{Components}}} 
& \emph{Source image} & \yes & \yes & \yes & \yes & \yes & \yes & \yes & \yes & \yes & \yes & \yes & \yes & \yes 
\\
& \emph{Target image} & \yes & \yes & \no & \yes & \yes & \yes & \no & \yes & \yes & \yes & \yes & \no & \no 
\\
& \emph{Fractal image} & \no & \no & \no & \no & \no & \no & \no & \yes & \no & \yes & \no & \yes & \yes 
\\
& \emph{Textual prompts} & \no & \no & \no & \no & \no & \no & \no & \no & \no & \no & \no & \yes & \yes 
\\
& \emph{Specific prompts} & \no & \no & \no & \no & \no & \no & \no & \no & \no & \no & \no & \no & \yes 
\\
& \emph{Interpolation} & \no & \yes & \yes & \no & \yes & \no & \no & \yes & \yes & \yes & \yes & \no & \yes 
\\
& \emph{Concatenation} & \yes & \no & \no & \yes & \yes & \yes & \yes & \yes & \no & \no & \no & \yes & \yes 
\\
\hline

\multirow{9}{*}{\rotatebox{90}{ \emph{Tasks} }}
& \emph{Adversarial robustness} & \yes & \yes & \yes & \yes & \yes & \yes & \yes & \yes & \yes & \yes & \yes & \yes & \yes 
\\
& \emph{General classification} & \yes & \yes & \yes & \yes & \yes & \yes & \yes & \yes & \yes & \yes & \yes & \yes & \yes 
\\
& \emph{Fine-grained} & \no & \no & \no & \no & \no & \no & \no & \no & \no & \no & \yes & \yes & \yes 
\\
& \emph{Transfer learning} & \yes & \no & \no & \yes & \no & \no & \no & \no & \no & \no & \no & \yes & \yes 
\\
& \emph{Data scarcity} & \no & \no & \no & \no & \no & \no & \no & \no & \no & \no & \yes & \yes & \yes 
\\
& \emph{Open-partial} & \no & \no & \no & \no & \no & \no & \no & \no & \no & \no & \no & \no & \yes \\
& \emph{Open-set} & \no & \no & \no & \no & \no & \no & \no & \no & \no & \no & \no & \no & \yes \\
& \emph{Partial-set} & \no & \no & \no & \no & \no & \no & \no & \no & \no & \no & \no & \no & \yes \\
\hline

\end{tabular}

\vspace{-1em}
\end{table*}

A preliminary version of this work was presented at a conference~\cite{islam2024diffusemix}. The present manuscript substantially extends that early effort by introducing several novel contributions. First, the DiffuseMix~\cite{moser2024diffusion} lacks mechanisms for reliability and responsible image editing, we address this gap by proposing an unsupervised data augmentation strategy. Our method extracts robust features from generative images without requiring labels, thereby filtering out undesirable synthetic samples. Second, we introduce a novel label-preserving approach, termed PatchSwap, which exchanges patches between generative and natural images. Third, we design a smoother concatenation technique that fuses original and synthetic images through a smoothness factor, consistently improving model performance. Fourth, we develop new contextual prompts tailored for domain adaptation scenarios including open-partial, open-set, and partial-set adaptation. We demonstrate that training the state-of-the-art LEAD framework~\cite{qu2024lead} with our proposed method yields remarkable gains across diverse categories. To the best of our knowledge, no existing approach jointly addresses both general image classification and domain adaptation tasks within a unified framework. Finally, we extend our empirical study to a wide range of architectures, spanning small-scale backbones (ResNet-18, ResNet-34, ResNet-50), large-scale backbones (ResNeXt-50, ResNet-101), and vision transformers (Swin Transformer, ConvNeXt), confirming the generality and scalability of our approach.




\section{Related Work}
\label{sec:relatedWork}

Data augmentation is widely adopted  to enhance the diversity of training datasets. Traditional methods apply basic transformations, such as flipping, rotation, and color jittering, to create new images from the original data. However, while these transformations can increase performance and improve generalization, they often lack semantic diversity, failing to capture key features critical for more complex tasks. This motivates the need for more advanced augmentation techniques that can introduce greater variation in the data while preserving essential content.

\par
\vspace{5pt}
\noindent{\textbf{Diffusion Models for Augmentation.}} 
Recently, there has been a growing interest in leveraging Diffusion Models (DMs) for data augmentation. Azizi et al. \cite{azizi2023synthetic} proposed using text-to-image fine-tuned diffusion models to generate synthetic samples for ImageNet classification, demonstrating that incorporating these generated images into the training set can enhance classification performance. Similarly, Trabucco et al. \cite{trabucco2023effective} utilized off-the-shelf diffusion models to create diverse and semantically rich images via prompt engineering, aiming to improve image classification on various datasets. Furthermore, Li et al. \cite{li2023synthetic} explored using diffusion models for data augmentation in conjunction with knowledge distillation, particularly in scenarios where real images are unavailable. These efforts underscore the potential of diffusion models to enrich data augmentation techniques.

\par
\vspace{5pt}
\noindent{\textbf{Image Mixing for Augmentation.}} 
Image mixing has become a widely employed  category of data augmentation techniques, enhancing the performance and robustness of deep learning models \cite{liang2023miamix, yan2023locmix, mensink2023infinite, chen2023rankmix}. Notable state-of-the-art methods in this category include CutMix, AugMix and PuzzleMix. CutMix \cite{yun2019cutmix} improves out-of-distribution generalization by randomly overlaying a patch from a source image onto a target image. AugMix~\cite{hendrycksaugmix} stochastically combines augmentation operations like equalize, posterize, and translate-x to introduce semantic diversity. 
PuzzleMix~\cite{kim2020puzzle} refines mixup techniques by considering image saliency and local statistics.
Similarly, another technique SaliencyMix~\cite{uddin2020saliencymix} leverages saliency maps to focus on the most important image regions, preserving their integrity. Manifold Mixup \cite{verma2019manifold} blends hidden network states during training, creating an interpolated state for improved feature representation. Another well-known method is PixMix \cite{hendrycks2022pixmix} that extends AugMix~\cite{hendrycksaugmix} by mixing input images with fractal and feature visualization images, improving \textcolor{blue}{machine learning safety \cite{yi2020improving}}. A comprehensive overview of various image-mixing methods, along with their components and task accomplishment is presented in Table \ref{table:compar_study}.
\par
\vspace{5pt}
\noindent{\textbf{Automated Data Augmentation.}} Several researchers have also explored automated data augmentation for improving model performance. For instance, AutoAugment \cite{cubuk2018autoaugment} utilizes reinforcement learning to discover optimal data augmentation strategies, while RandAugment \cite{cubuk2020randaugment} employs a set of randomized augmentation operations to enhance model generalization and robustness. AdaAug \cite{cheung2021adaaug} has been  introduced to efficiently learn adaptive augmentation policies that vary depending on the class or even the instance.
\par 

Although existing methods are effective within their settings, they often lack in generalizability. The conventional image mixing techniques remain brittle due to their non-adaptive nature while more recent generative modeling based methods underperform due to poorly crafted prompts. Moreover, the desired augmentation image formation by these methods largely ignore the semantics while blending the available images. 
Unlike these previous techniques, our method focuses on combining original and generated images in a more sophisticated manner for semantic preservation while also leveraging a predefined library of conditional prompts. Our hybrid images are also blended with fractal patterns to further boost the overall model generalization and robustness.

\begin{figure}
    \centering
    \includegraphics[width=0.49\textwidth]{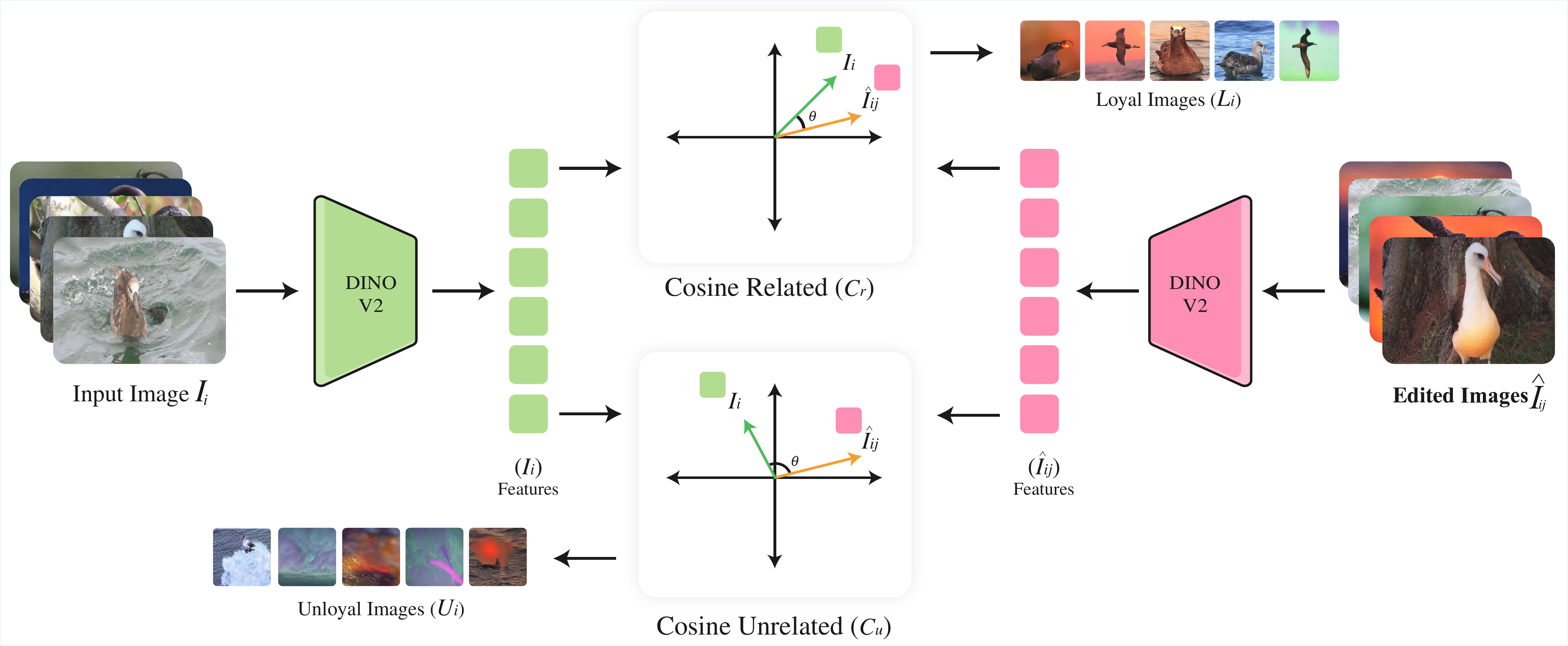}
    \caption{ Overview of the proposed filtration module without label supervision.}
    \label{fig:filtration_GMix}
    \vspace{-10pt}
\end{figure}

\begin{figure*}[t]
    \centering
    \includegraphics[width=0.98\textwidth]{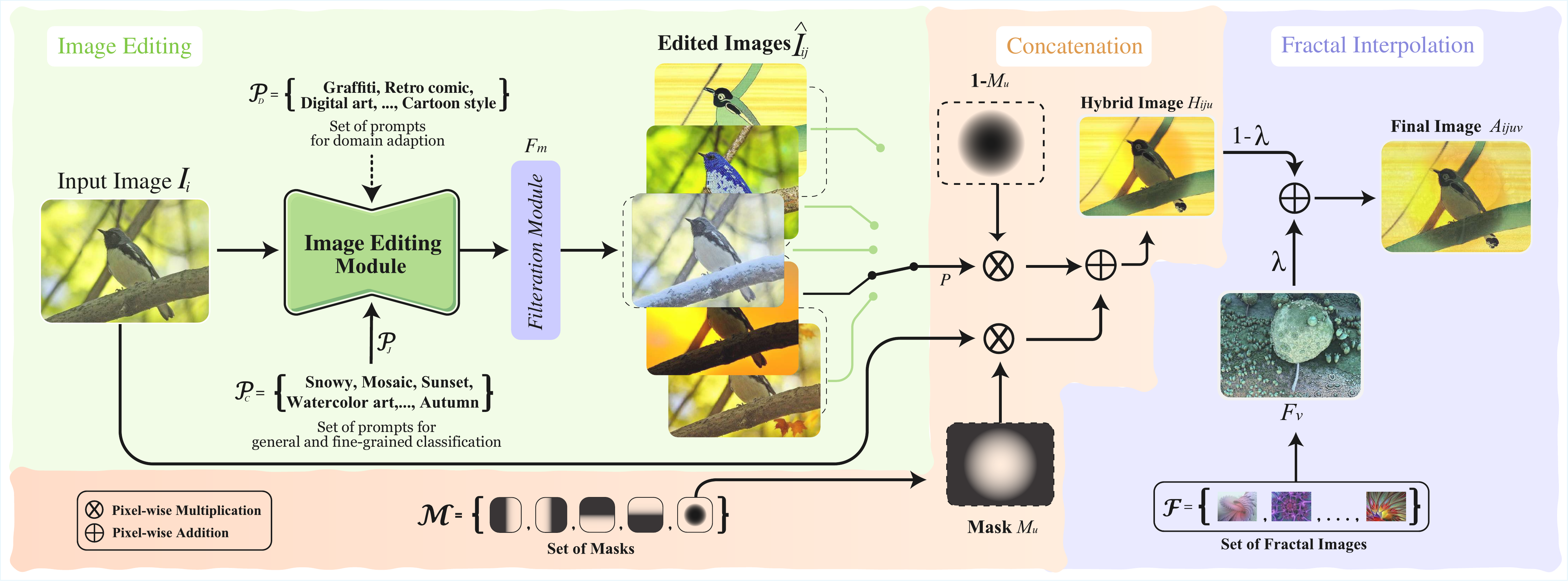}
    \caption{ \textbf{Overview of the proposed \our{} method}. An input image and a randomly selected prompt are input to a pre-trained image editing diffusion model to obtain a generated image. Input and generated images are smoothly concatenated using a binary mask to obtain a hybrid image. Random fractal image is blended with this hybrid image to obtain the augmented image.}
    \label{fig:GMixPipeline}
    \vspace{-10pt}
\end{figure*}

\section{\textsc{Proposed Method}}
\label{sec:method}

\subsection{Overview}
Existing image mixing based data augmentation techniques may induce label ambiguities by pasting the source image contents onto  the target image and consequently replacing either important regions of the foreground  or the required background context  \cite{cutmix, kang2023guidedmixup}. The central idea of our approach is to augment the source image only while preserving the key semantics and  concatenating a portion of the source image with its generative counterpart.  At the same time, our method introduces diverse object details and contexts in the resulting image for effective augmentation.

The schematics of the proposed \our{} technique is provided in Fig.~\ref{fig:GMixPipeline}.
Our method comprises three main steps: 
\textbf{\emph{prompt-guided image editing}, \emph{seamless concatenation}, \emph{selection of faithful images}}, and \textbf{ \emph{fractal interpolation}}. 
Initially, we use conditional prompts with a diffusion-based image editing module to compute generative counterparts of the source image. Then, a portion of the source image is concatenated with the rest of the generated image, forming a hybrid image. This approach ensures that the training model has access to the original visual content along with the generated one. Finally, the hybrid image is blended with a random fractal image, obtaining the final training image with a diverse structure. Improvements of our image by blending fractal is inspired by~\cite{hendrycks2022pixmix, hendrycks2021unsolved}. However, in our case, the blending objective for augmentation is unique to our approach. 




\subsection{GenMix Details}

The proposed \our{} is quite effective data augmentation method that aims to enhance the generalization and robustness of the deep neural networks. 
Formally,   $I_i \in \mathbb{R}^{h\times w \times c}$ is a training dataset image.  Our data augmentation method is defined as: $\mathcal{D}$\textsubscript{mix}$(\cdot): \mathbb{R}^{h\times w \times c} \to \mathbb{R}^{h\times w \times c}$. 
Input image $I_i$ goes through proposed generation using randomly selected  prompt $p_j$, seamless concatenation using random mask $M_u$, and linearly interpolated using a random fractal image $F_v$ to obtain the final augmented image $A_{ijuv}$. The overall augmentation process, as shown in Algorithm \ref{algo:diffuseMix}, may be represented as
\begin{equation}
    A_{ijuv} = \mathcal{D}_{\text{mix}}(I_i, p_j, M_u, F_v, \lambda).
    \label{equ:first}
\end{equation}



\begin{figure*}[t]
\centering
\includegraphics[width=0.90\textwidth]{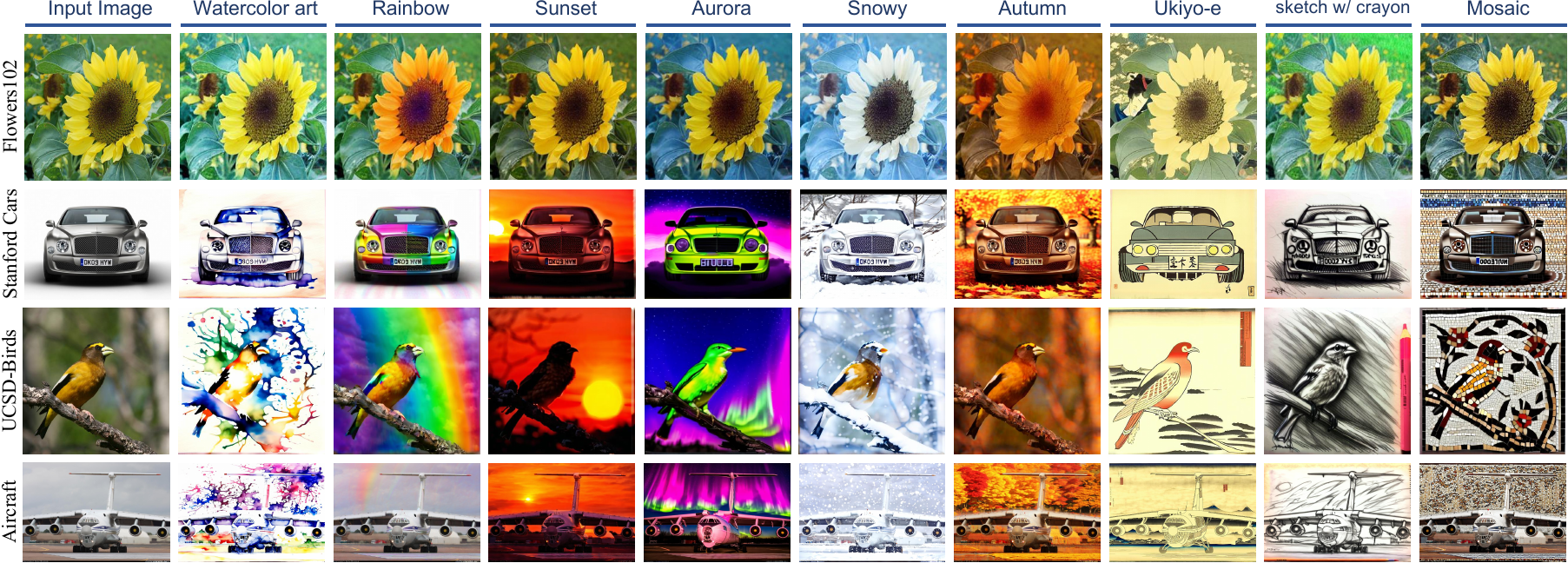}
\caption{Different \emph{bespoke conditional prompts} - columns -  are used to obtain generated images preserving important features and adding rich visual appearance to the input images. Each row shows a representative input from a different dataset.}
\label{fig:prompts_variation}
\end{figure*}

\vspace{5pt}
\noindent \textbf{Prompt Guided Image Editing:}
Our image editing step $\mathcal{E}(.)$ consists of a \textit{pre-trained} image editing diffusion model that requires a prompt $p_j$ from a predefined set of $k$ prompts, $P = \{p_1, p_2, \dots, p_k\}$ where $j \in [1, k]$ along with the input training image $I_i$ and produces an edited counterpart image $\hat{I}_{ij}$. The image generation process in conventional diffusion models is often open-ended for image-to-image or text-to-image translation. On the other hand, image editing is guided by textual prompts.  The aim  is to obtain diverse image-to-image translations guided by textual prompts.
In our case, as the goal is to achieve a slightly modified but not too different version of $I_i$, \emph{filter-like} prompts are curated in $\mathcal{P}$ which do not alter the image drastically. Examples of the prompts used in \our{} are shown in Figure \ref{fig:prompts_variation}.
The overall generation step can be represented as: 
\begin{equation}
    \hat{I}_{ij} = \mathcal{G}(I_i, p_j), \text{ where } p_j \in \mathcal{P}
\end{equation}

\vspace{5pt}

\noindent \textbf{Selection of Faithful Images:} To ensure semantic consistency, we retain only those generated images whose similarity to their corresponding original image exceeds a dynamic threshold. Formally, the selection criterion for faithful images is defined as:
\begin{equation}
\hat{I}_{ij} \in \mathcal{F} \quad \text{if} \quad \text{cos\_sim}\left(f_m(I_i), f_m(\hat{I}_{ij})\right) \geq \mu - 2\sigma
\label{eq:faithful}
\end{equation}

\noindent
where $f_m(\cdot)$ is the DINOv2 feature extractor, $\mu$ and $\sigma$ denote the mean and standard deviation of pairwise cosine similarities computed among all original images $I_i$, and $\mathcal{F}$ denotes the set of faithful generated images retained for augmentation. As defined in Eq.~\eqref{eq:faithful}, we discard dissimilar generations and retain only faithful images for downstream augmentation.

\noindent \textbf{Seamless Concatenation:} We concatenate a portion of the original input image $I_i$ with its edited version $\hat{I}_{ij}$  using a randomly selected mask $M_u$ from the set of masks  to create a hybrid image $H_{iju}$: 
\begin{equation}
    H_{iju}= (\hat{I}_{ij} \odot M_u) + (I_{i} \odot (\bold{1}-M_u)),
    \label{equ:one}
\end{equation}
where $M_u$ is a mask, $\odot$ is a pixel-wise multiplication operator. The set of masks contains four kinds of masks including horizontal, vertical and flipped versions.  Such masking ensures the availability of the semantics of the input image to the learning network while reaping the benefits of the generated images.

The mask $M_u$ consists of zeros, ones, and smoothly varying values for seamless concatenation. To create a smooth transition between two images, our mask consists of three parts: $M_0 \in \mathbb{R}^{h\times (w-b)/2}$ containing zeros, $M_1 \in \mathbb{R}^{h\times (w-b)/2}$ containing ones, and $M_b \in \mathbb{R}^{h\times b}$ containing the blending matrix where $b$ is the blending width.  
These three parts are stacked to form the  mask:
$M_{u} = \begin{bmatrix} M_{0} & M_b & M_{1} \end{bmatrix}$ either horizontally or vertically.
We generate the blending matrix $M_b$ such that the values smoothly transition from 0 on the $M_0$ side to 1 on the $M_1$ side, over the specified blending width $b$. The same $M_u$ is applied across all three color channels of one image and (1-$M_u$) is applied to the second image and then both images are added to get the hybrid image as shown by Eq. \eqref{equ:one}.

\begin{table*}[t]
    \centering
    \caption{ Top-1 (\%) performance of mixup methods on CIFAR-100, Tiny-ImageNet and ImageNet-1K. The previous SOTA miup methods results are taken from AdAutoMix \cite{qin2023adversarial} and reproduce DiffuseMix \cite{islam2024diffusemix} results on new backbones.}
    \vspace{-5pt}

    \begin{tabular}{l|cc|cc|cc|ccc} \toprule
            & \multicolumn{2}{c|}{\textbf{CIFAR100}}    & \multicolumn{2}{c|}{\textbf{CIFAR100}} & \multicolumn{2}{c|}{\textbf{Tiny-ImageNet}}  & \multicolumn{3}{c}{\textbf{ImageNet-1K}} 
            \\
  \textbf{Method}          & \textbf{ResNet18}      & \textbf{ResNeXt50}         & \textbf{Swin-T}         & \textbf{ConvNeXt-T}  & \textbf{ResNet18}      & \textbf{ResNeXt50}  & \textbf{ResNet18}   & \textbf{ResNet34}    & \textbf{ResNet50} 
  \\ 
  \midrule
    Vanilla \cite{resnet}         & 78.04         & 81.09             & 78.41          & 78.70       & 61.68         & 65.04      & 70.04      & 73.85      & 76.83       \\
    MixUp \cite{mixup}           & 79.12         & 82.10             & 76.78          & 81.13       & 63.86         & 66.36      & 69.98      & 73.97      & 77.12       \\
    CutMix \cite{cutmix}         & 78.17         & 81.67             & 80.64          & 82.46       & 65.53         & 66.47      & 68.95      & 73.58      & 77.17       \\
    SaliencyMix \cite{uddin2020saliencymix}     & 79.12         & 81.53             & 80.40          & 82.82       & 64.60         & 66.55      & 69.16      & 73.56      & 77.14       \\
    FMix \cite{harris2020fmix}           & 79.69         & 81.90             & 80.72          & 81.79       & 63.47         & 65.08      & 69.96      & 74.08      & 77.19       \\
    PuzzleMix \cite{kim2020puzzle}       & 81.13         & 82.85             & 80.33          & 82.29       & 65.81         & 67.83      & 70.12      & 74.26      & 77.54       \\
    ResizeMix \cite{qin2020resizemix}       & 80.01         & 81.82             & 80.16          & 82.53       & 63.74         & 65.87      & 69.50      & 73.88      & 77.42       \\
    AutoMix \cite{liu2022automix}         & 82.04   & 83.64       & 82.67    & 83.30 &   67.33   &   70.72 &  70.50 &  74.52 &  77.91 \\
    AdAutoMix \cite{qin2023adversarial} & 82.32 & 84.22 & 84.33 & 83.54 & 69.19 & 72.89 & 70.86 & 74.82 & 78.04   
    \\ 
        \midrule
    DiffuseMix \cite{islam2024diffusemix}    & 82.78     &  84.88       & 85.26      & 84.38   & 70.47    & 73.62  & 71.62  & 75.23      & 78.64  
    \\ 

    \bf{\our{}}    &\bf{84.25}     &\bf{86.23}       &\bf{86.28}      &\bf{85.25}   &\bf{71.38}     &\bf{74.81}  &\bf{72.32}  &\bf{75.14}      &\bf{79.32} \\   \bottomrule
    \end{tabular}
    
    \label{tab:supp_combined}
\end{table*}

\begin{algorithm}
\caption{ \our{}}
\begin{algorithmic}[1]
\Require $I_i \in \mathcal{D}$ training images dataset, $m$: number of augmented images,
 $p_j \in \mathcal{P}$ set of textual prompts, $M_u \in \mathcal{M}$ set of smooth masks, $F_v \in \mathcal{F}$ set of self-similarity fractal images, $\lambda$: interpolation weight 
\Ensure $\mathcal{D}'$: $m$ Augmented images  
\State $\mathcal{D}' \gets \emptyset$ 
\For{each image $I_i$ in $\mathcal{D}$}
\For{$a$ in  $\{1:m\}$}
    \State Randomly select prompt $p_j$ from $\mathcal{P}$ 
    \State Image Editing: $\hat{I}_{ij} \gets \mathcal{E}(I_i, p_j)$ 
       \State Randomly select mask $M_u$ from $\mathcal{M}$
            \State Concatenation: $H_{iju} \gets$  $M_u \odot I_i+(1-M_u) \odot \hat{I}_{ij}$ 
            \State  Randomly select $F_v$ from $\mathcal{F}$
            \State Linear Interpolation: $A_{ijuv} \gets (1-\lambda) H_{iju}+ \lambda F_v$ 
            \State Add $A_{ijuv}$ to $\mathcal{D}'$
    \EndFor
\EndFor

\State \textbf{return} $D'$
\end{algorithmic}

\label{algo:diffuseMix}
\end{algorithm}

\vspace{5pt}
\noindent \textbf{Fractal Interpolation:}
A self-similarity fractals dataset $\mathcal{F}$ is collected and used for inducing structural variations in the hybrid images.
The interpolated image $H_{iju}$ is linearly computed between a randomly  selected fractal image $F_v \in \mathcal{F}$ and the hybrid image $H_{iju}$:
\begin{equation}
    A_{ijuv}= \lambda F_v + (1-\lambda) H_{iju}, \text{  }  0 \le \lambda < 1
     \label{equ:two}
\end{equation}

\noindent where $\lambda$ is the interpolation factor. This results in the final augmented image $A_{ijuv}$  used to train or fine-tune deep neural networks. The overall augmentation process of \our{} can be represented as:
\begin{equation}
    A_{ijuv} =(1 - \lambda)(I_i \odot M_u + \hat{I}_{ij} \odot (\bold{1}-M_u))+\lambda F_v,
    \label{eq:masking}
\end{equation}
An ablation study is conducted to find the appropriate value of $\lambda$.

\section{Experimental Setup}
\label{sec:experiment}
\par
\vspace{2pt}
\noindent{\textbf{Datasets.}} To enable comparisons with previous research on image augmentation \cite{cutmix, mixup, kang2023guidedmixup, uddin2020saliencymix, hendrycksaugmix, cheung2021adaaug, cubuk2020randaugment, autoaugment, kim2020puzzle, kim2020co, huang2021snapmix, lim2019fast, verma2019manifold}, we assess our approach using both \emph{general image classification} and \emph{fine-grained image classification} datasets. For general image classification, we use three datasets: {ImageNet-1K} 
\cite{imagenet}, CIFAR100 \cite{krizhevsky2009learning}, and Tiny-ImageNet-200 \cite{le2015tiny}. For fine-grained image classification, we employ three datasets: CUB \cite{nilsback2008automated}, Stanford Cars \cite{krause20133d}, and Aircraft~\cite{maji2013fine}. 
Moreover,  we use Flower102~\cite{nilsback2008automated}, Aircraft~\cite{maji2013fine},
and Stanford Cars~\cite{krause20133d} for evaluation related to Transfer Learning. The data scarcity related experiments are conducted with Flower102~\cite{nilsback2008automated} dataset. Our experiments for domain adaption additionally use OfficeHome \cite{venkateswara2017deep} dataset. 
These datasets encompass a variety of scenarios, featuring images with a broad spectrum of objects like plants, animals, vehicles, human actions, and general objects across diverse scenes and textures.  

\vspace{2pt}
\noindent{\textbf{Implementation Details.}} For image editing, we employed InstructPix2Pix pre-trained image editing diffusion model \cite{brooks2023instructpix2pix} utilizing our custom-built textual prompts. To create the Mask $M_u$ we used a fixed blending width of 20 pixels. We use $\lambda=0.20$ for blending the fractal image in  Eq. \eqref{equ:two}.  

\vspace{3pt}
\noindent{\textbf{Textual Prompt Selection.}} To ensure that only suitable prompts are used for classification, a custom textual library of \textit{filter-like} global visual effects is predefined, comprising  \textcolor{customviolet}{\emph{`autumn', `snowy', `sunset', `watercolor art', `rainbow', `aurora', `mosaic', `ukiyo-e'}} and  \textcolor{customviolet}{\emph{`a sketch with crayon'}}. For domain adaptation, we introduce an additional set of bespoke conditional prompts, such as \textcolor{customviolet}{\emph{`graffiti', `retro comic', `chalk drawing', `watercolor painting', `digital art'}} and \textcolor{customviolet}{\emph{`cartoon style'}}. These prompts are designed to introduce a broader range of visual appearances, enhancing the model's adaptability to different domains ensuring their generic nature and wide applicability. These prompts do not significantly edit the image structure while introducing  global visual effects. Each prompt in the textual library is appended with the template `A transformed version of image into $prompt$' to form a specific input for the image editing diffusion model. Examples of images generated using these prompts are shown in Figure~\ref{fig:prompts_variation}.

\begin{table*}[t]
    \centering
    \caption{FGSM error rates on CIFAR-100 and Tiny-ImageNet-200 datasets for PreActResNet-18, following the protocols previous SOTA mixup methods \cite{islam2024diffusemix, kang2023guidedmixup} and we reproduce results for AdAutoMix \cite{qin2023adversarial}. }
    \vspace{-5pt}
    \resizebox{0.85\linewidth}{!}{
    \begin{tabular}{l|ccccccc|cc} \toprule
     \textbf{Dataset}  & Vanilla & Mixup   & Manifold  & CutMix & AugMix  & PuzzleMix & AdAutoMix & DiffuseMix & \textbf{\our{}}  
     \\ 
     \midrule
     CIFAR-100  & 23.67  & 23.16      & 20.98         & 23.20        & 43.33 & 19.62 &  24.34 & 17.38 & \textbf{16.83}     \\
     Tiny-ImageNet-200   & 42.77   & 43.41      & 41.99         & 43.33        & --   & 36.52 & -- & 34.53 & \textbf{33.83}  \\ \bottomrule
     \end{tabular}
      }
    \label{tab:fgsm}
\end{table*}

\section{Performance Evaluation}

\subsection{General Classification }
General Classification (GC) is an essential task used to assess the effectiveness of data augmentation approaches. An increase in GC accuracy indicates that the augmentation method fairly introduces meaningful variations in input data to enhance the learning process. In our evaluation, we applied our method to three challenging datasets: CIFAR-100 \cite{krizhevsky2009learning}, ImageNet-1K \cite{imagenet}, and Tiny-ImageNet-200 \cite{le2015tiny}. Aligning with the practices of current state-of-the-art (SOTA) methods, we used ResNet-50 for the ImageNet-1K, and utilized ResNet-18 \cite{he2016deep} and ResNeXt-50 \cite{resnext} for the Tiny-ImageNet-200 and CIFAR-100 datasets \cite{huang2021snapmix, kang2023guidedmixup, kim2020co, uddin2020saliencymix}, and employed ResNet-50 for the ImageNet dataset \cite{kim2020puzzle, hendrycksaugmix, cutmix}.
\par
Table \ref{tab:supp_combined} showcases a comparison of Top-1 performances on the ImageNet-1K dataset and compares with the existing SOTA methods. \our{} exhibited strong performance, achieving a Top-1 accuracy gain of \numbersBlue{$2.49\%$} over the Vanilla \cite{resnet} baseline. Compared to DiffuseMix \cite{islam2024diffusemix}, which shows the second-best Top-1 result of $78.64\%$, our approach attains an absolute performance improvement of \numbersBlue{$0.68\%$}. Table \ref{tab:supp_combined} also showcases a comparison of performances on Tiny-ImageNet-200 with existing SOTA methods. Our proposed \our{} demonstrates superior performance over existing data augmentation approaches. Over the Vanilla \cite{resnet}, we obtained an absolute Top-1 performance gain of \numbersBlue{$9.77\%$}, and an improvement of \numbersBlue{$1.19\%$} over DiffuseMix \cite{islam2024diffusemix}.
\par
We observe similar trends on the CIFAR-100 dataset. \our{} outperforms the Vanilla \cite{resnet} with a Top-1 absolute accuracy gain of \numbersBlue{$6.21\%$} using ResNet-18. Compared to DiffuseMix \cite{islam2024diffusemix}, our method achieves a gain of \numbersBlue{$1.47\%$} on the same backbone. These results across diverse and challenging benchmark datasets underscore the effectiveness of \our{} in enhancing learning outcomes. They also indicate its capability to mitigate overfitting and promote better generalization in neural network training.

\subsection{Adversarial Robustness}
In line with current state-of-the-art methods \cite{kim2020puzzle,cutmix,hendrycksaugmix, verma2019manifold,uddin2020saliencymix,hong2021stylemix,kim2020co,kang2023guidedmixup, pan2024enhanced}, we assess the robustness of \our{} method against adversarial attacks and input perturbations for general classification. For these evaluations, we employ fast adversarial training \cite{fast_imagenet} to generate adversarially perturbed input images. The primary objective of these tests is to determine if our augmentation technique offers enhanced resistance to adversarial attacks. We measure performance using the FGSM \cite{fast_imagenet} error rates as an indicator of robustness against such attacks.
\par
As presented in Table \ref{tab:fgsm}, our approach achieves an FGSM error rate of \numbersBlue{$16.83\%$} on CIFAR-100, outperforming all other methods in comparison. Similarly, on Tiny ImageNet-200, our method outperforms existing state-of-the-art techniques with an error rate of \numbersBlue{$33.83\%$}. These results highlight that our method maintains strong resilience to adversarial perturbations, surpassing the performance of existing top-performing approaches.

\begin{table}

\caption{ Top-1 (\%) performance comparison for fined-grained tasks. Previous results are taken from AdAutoMix \cite{qin2023adversarial} and reproduce DiffuseMix \cite{islam2024diffusemix} results on new backbones. }
\vspace{-5pt}
\label{tab:fgvc_cars_aircraft}
\centering
\begin{tabular}{l|cc|cc}
\toprule
\multirow{2}{*}{ \textbf{Method}} & \multicolumn{2}{c|}{\textbf{FGVC-Aircrafts}} & \multicolumn{2}{c}{\textbf{Stanford-Cars}} \\
                        & \textbf{ResNet18}  & \textbf{ResNeXt50} & \textbf{ResNet18} & \textbf{ResNeXt50} \\
\midrule
Vanilla~\cite{resnet}              & 80.23 & 85.10 & 86.32 & 90.15 \\
Mixup~\cite{mixup}                 & 79.52 & 85.18 & 86.27 & 90.81 \\
CutMix~\cite{cutmix}               & 78.84 & 84.55 & 87.48 & 91.22 \\
ManiMix~\cite{verma2019manifold} & 80.68 & 86.60 & 85.88 & 90.20 \\
SalieMix~\cite{uddin2020saliencymix} & 80.02 & 84.31 & 85.88 & 90.20 \\
FMix~\cite{harris2020fmix}   & 79.36 & 86.23 & 87.55 & 90.90 \\
PuzzleMix~\cite{huang2021snapmix}       & 80.76 & 86.23 & 87.78 & 91.29 \\
ResizeMix~\cite{kim2020puzzle}        & 78.10 & 84.08 & 88.17 & 91.36 \\
AutoMix~\cite{kim2020co}             & 81.37 & 86.72 & 88.89 & 91.38 \\
AdAuto\cite{kang2023guidedmixup} & 81.73  &  87.16  & 89.19 & 91.59  \\
\midrule
DiffuseMix\cite{islam2024diffusemix}  & 82.69 & 88.13 & 90.25 & 92.36 \\
\textbf{\our{}} & \textbf{83.87} & \textbf{89.27} & \textbf{91.54} & \textbf{93.21} \\
\bottomrule
\end{tabular}
\end{table}

\begin{table*}[t]
    \centering
    \caption{Top-1 (\%) accuracy on the data scarcity experiment under the setting ($10$ images per class).}
    \vspace{-5pt}
    \resizebox{0.98\linewidth}{!}{
    \begin{tabular}{l|cccccccc|cc}
    \toprule
    \textbf{Method} & Vanilla & Mixup & CutMix & SaliencyMix & SnapMix &  PuzzleMix & Co-Mixup  & GuidedMixup & DiffuseMix & \textbf{\our{}} \\
    \midrule
    Validation & 64.48 & 70.55 & 62.68 & 63.23 & 65.71 & 71.56 & 68.17 &74.74 & 77.14 & \textbf{79.34}  \\
    Test & 59.14 & 66.81 & 58.51 & 57.45 & 59.79 & 66.71 & 63.20 & 70.44 &  74.12 & \textbf{ 76.38} \\
    \bottomrule
    \end{tabular}
    }
    \label{tab:data_scarcity}
\end{table*}


\begin{table*}[t]
    \centering
    \caption{Top-1 (\%) performance on ResNet50 with CUB200 and Standford-Cars datasets.}
    \vspace{-5pt}
    \resizebox{0.85\linewidth}{!}{
    \begin{tabular}{l|cccccc|cc}
    \toprule
    \textbf{Method} & Vanilla & Mixup & CutMix & PuzzleMix & AutoMix & AdAutoMix & DiffuseMix & \textbf{\our{}} \\
    \midrule
    CUB-Birds 200 & 81.76 & 82.79 & 81.67 & 82.59 & 82.93 & 83.36 & 83.83 & \textbf{84.45}  \\
    Stanford-Cars  & 88.88 & 89.45 & 88.99 & 89.37 & 88.71 & 89.65 & 90.26 & \textbf{90.84} \\
    \bottomrule
    \end{tabular}
    }
    \label{tab:transfer_learning}
    \vspace{-10pt}
\end{table*}

\subsection{Fine-Grained Visual Classification}
Fine-Grained Visual Classification (FGVC) poses arduous challenge compared to general classification because it requires distinguishing subtle variations between objects within the similar category with broader range \cite{pu2024fine}. To effectively train a model on such tasks, a generative augmentation approach must carefully preserve these fine-grained details. We assess the performance of \our{} on FGVC task using two well-known datasets: Stanford Cars \cite{krause20133d}, and FGVC-Aircraft~\cite{maji2013fine}, employing the ResNet-50 \cite{resnet} and ResNeXt \cite{xie2017aggregated} architectures. 

As demonstrated in Table \ref{tab:fgvc_cars_aircraft}, \our{} outperforms other competitive methods, showing its strength in enhancing the model's ability to generalize. It is notable that \our{} achieves the highest accuracy across all the datasets. On the FGVC-Aircraft dataset, it achieves an absolute gain of top  \numbersBlue{$1.18\%$} on ResNet18 and \numbersBlue{$1.14\%$} on ResNext50 over the strong baseline of DiffuseMix ~\cite{islam2024diffusemix}. For the Stanford-Cars dataset, \our{} achieves an absolute gain of \numbersBlue{$1.29\%$} on ResNet18 and \numbersBlue{$0.85\%$} on ResNext50. Lastly, on the Stanford Cars dataset, the absolute gain is again more than ~1\%. This consistent performance gain of \our{} can be attributed to the preservation of subtle visual information in the hybrid image constructed in our approach.

\subsection{Transfer Learning}

Transfer learning is a prevalent technique for adapting large neural network architectures  pre-trained on large-scale datasets to smaller datasets.  It requires limited computational resources and helps conducting experiments rapidly. Notably, many augmentation methods \cite{cutmix,mixup,kang2023guidedmixup,uddin2020saliencymix,hendrycksaugmix,kim2020puzzle,kim2020co,huang2021snapmix,verma2019manifold} have not reported their performance for this popular application. Nevertheless, we evaluate our approach by fine-tuning the baseline model on three datasets Flower102, Aircraft, and Stanford Cars, utilizing ResNet-50 model pre-trained on ImageNet. The results are summarized in Table~\ref{tab:transfer_learning}.

On the CUB-Birds-200, \our{} achieved the highest accuracy of \numbersBlue{$84.45\%$}. A similar trend is evident in the Stanford-Cars dataset, where \our{} reached \numbersBlue{$90.84\%$}. \our{} significantly surpassed other augmentation techniques, demonstrating its effectiveness in transfer learning scenarios. As compared to the vanilla baselines, our method achieves absolute performance gains of \numbersBlue{$2.69\%$}, and \numbersBlue{$1.96\%$} on the used datasets. Since fine-tuning requires substantially fewer computational resources than training from scratch, this experiment highlights the practical importance of  \our{} in deep learning applications.

\subsection{Data Scarcity}
Data scarcity presents a significant challenge when training deep neural networks. With only a limited number of examples per class, deep networks may struggle to learn meaningful patterns, which can lead to overfitting and a reduced ability to generalize \cite{zhang2024few, chen2019multi}. To mitigate this, augmentation techniques are commonly applied to increase the amount of training data. In this context, we evaluate the performance of ResNet-18 \cite{resnet} trained with just $10$ images per class from the original Flower102 dataset. As shown in Table \ref{tab:data_scarcity}, \our{}  consistently outperforms other mixup approaches, achieving Top-1 validation accuracy is \numbersBlue{$79.34\%$} and Top-5 accuracy of \numbersBlue{$76.38\%$}. Our method is specifically designed to diversify the training dataset. By using custom conditional prompts, \our{} artificially expands and enriches the dataset, leading to stronger neural network learning.

\begin{table*}
  \centering
\caption{Comparison of H-score (\%) in the Open Partial Domain Adaptation (OPDA) scenario on the Office-Home, where Clipart (Cl), Art (Ar), Real (Re), and Product (Pr) are different domains. Unified (U) methods are applicable for all  scenarios
including OPDA, OSDA, and PDA. The SF denotes source data-free, and CF indicates K-means clustering free model adaption. Our method is shown as +{\our}  to imply LEAD+\our{}. Boldface represents performance gains of \our{} over LEAD.} 
\vspace{-5pt}


  \addtolength{\tabcolsep}{0.0pt}
  \resizebox{1.0\textwidth}{!}{
    \begin{tabular}{lccc|ccccccccccccl}
    \toprule
     \textbf{Method} & \textbf{U} & \textbf{SF} & \textbf{CF} & \textbf{Ar2Cl} & \textbf{Ar2Pr} & \textbf{Ar2Re} & \textbf{Cl2Ar} & \textbf{Cl2Pr} & \textbf{Cl2Re} & \textbf{Pr2Ar} & \textbf{Pr2Cl} & \textbf{Pr2Re} & \textbf{Re2Ar} & \textbf{Re2Cl} & \textbf{Re2Pr} & \textbf{Avg} \\

    \midrule
    CMU~\cite{fu2020_cmu} & \xmark  & \xmark & \cmark &  56.0  & 56.9  & 59.2  & 67.0  & 64.3  & 67.8  & 54.7  & 51.1  & 66.4  & 68.2  & 57.9  & 69.7  & 61.6  \\
    DANCE~\cite{saito2020_unida_dance} & \cmark & \xmark & \cmark & 61.0  & 60.4  & 64.9  & 65.7  & 58.8  & 61.8  & 73.1  & 61.2  & 66.6  & 67.7  & 62.4  & 63.7  & 63.9  \\
    DCC~\cite{li2021_dcc} & \cmark & \xmark  &\xmark & 58.0  & 54.1  & 58.0  & 74.6  & 70.6  & 77.5  & 64.3  & {73.6}  & 74.9  & {81.0} & {75.1} & 80.4  & 70.2  \\
    OVANet~\cite{saito2021_ovanet} & \xmark & \xmark &\cmark & 62.8  & 75.6  & 78.6  & 70.7  & 68.8  & 75.0  & 71.3  & 58.6  & 80.5  & 76.1  & 64.1  & 78.9  & 71.8  \\
    GATE~\cite{chen2022_gate}  & \cmark & \xmark &\cmark & 63.8  & {75.9}  & {81.4}  & {74.0}  & {72.1}  & 79.8  & 74.7  & {70.3}  & {82.7} & 79.1  & 71.5  & {81.7}  & 75.6  \\
    UniOT~\cite{chang2022_uniot} & \xmark &\xmark &\cmark & 67.3 & 80.5 & 86.0 & 73.5 & 77.3 & 84.3 & 75.5 & 63.3 & 86.0 & 77.8 & 65.4 & 81.9 & {76.6} \\
    Source-only & - &- &- & 47.3  & 71.6  & 81.9  & 51.5  & 57.2  & 69.4  & 56.0  & 40.3  & 76.6  & 61.4  & 44.2  & 73.5  & 60.9  \\
    SHOT-O~\cite{liang2020_shot} & \xmark & \cmark & \cmark & 32.9  & 29.5  & 39.6  & 56.8  & 30.1  & 41.1  & 54.9  & 35.4  & 42.3  & 58.5  & 33.5  & 33.3  & 40.7  \\
    \midrule
    LEAD \cite{qu2024lead} & \cmark &\cmark & \cmark  & 62.7 & 78.1 & 86.4 & 70.6 & 76.3 & 83.4 & 75.3 & 60.6 & 86.2 & 75.4 & 60.7 & 83.7 & 75.0\\
    + \textbf{\our{}}   & \cmark &\cmark & \cmark  & \textbf{63.9} & \textbf{83.0} & 85.9 & \textbf{76.8}  & \textbf{78.9} & \textbf{85.3} & \textbf{78.1}  & \textbf{62.2}  & \textbf{86.9} & \textbf{79.8} & \textbf{62.9} & 83.0 & \textbf{77.2}\\

    \bottomrule
    \end{tabular}
    }
  \label{tab:opda_officehome}%
  
\end{table*}%

\begin{table*}
  \centering

\caption{H-score (\%) comparison in the Open-set Domain Adaptation (OSDA) scenario on the Office-Home dataset.} 
\vspace{-5pt}

  \resizebox{1.0\textwidth}{!}{
    \begin{tabular}{lccc|ccccccccccccc}
    \toprule
\textbf{Method} & \textbf{U} & \textbf{SF} & \textbf{CF} & \textbf{Ar2Cl} & \textbf{Ar2Pr} & \textbf{Ar2Re} & \textbf{Cl2Ar} & \textbf{Cl2Pr} & \textbf{Cl2Re} & \textbf{Pr2Ar} & \textbf{Pr2Cl} & \textbf{Pr2Re} & \textbf{Re2Ar} & \textbf{Re2Cl} & \textbf{Re2Pr} & \textbf{Avg} \\

    \midrule
    CMU~\cite{fu2020_cmu} & \xmark  & \xmark & \cmark & 55.0  & 57.0  & 59.0  & 59.3  & 58.2  & 60.6  & 59.2  & 51.3  & 61.2  & 61.9  & 53.5  & 55.3  & 57.6 \\
    DANCE~\cite{saito2020_unida_dance} & \cmark & \xmark & \cmark & 6.5   & 9.0   & 9.9   & 20.4  & 10.4  & 9.2   & 28.4  & 12.8  & 12.6  & 14.2  & 7.9   & 13.2  & 12.9 \\
    DCC~\cite{li2021_dcc} & \cmark  & \xmark  & \xmark & 56.1  & 67.5  & 66.7  & 49.6  & 66.5  & 64.0  & 55.8  & 53.0  & 70.5  & 61.6  & 57.2  & 71.9  & 61.7 \\
    OVANet~\cite{saito2021_ovanet} & \xmark & \xmark & \cmark & 58.6  & 66.3  & 69.9  & 62.0  & 65.2  & 68.6  & 59.8  & 53.4  & 69.3  & 68.7  & 59.6  & 66.7  & 64.0 \\
    Source-only &- &- &- & 46.1  & 63.3  & 72.9  & 42.8  & 54.0  & 58.7  & 47.8  & {36.1} & 66.2  & 60.8  & 45.3  & 68.2  & 55.2 
    \\
    SHOT-O~\cite{liang2020_shot} & \xmark & \cmark & \cmark & 37.7  & 41.8  & 48.4  & 56.4  & 39.8  & 40.9  & 60.0  & 41.5  & 49.7  & 61.8  & 41.4  & 43.6  & 46.9 
    \\ \midrule
     LEAD \cite{qu2024lead} & \cmark & \cmark & \cmark & 60.7 & 70.8 & 76.5 & 61.0 & 68.6 & 70.8 & 65.5 & 59.8 & 74.2 & 64.8 & 57.7 & 75.8 & 67.2
     \\
     + \textbf{\our{}}  & \cmark & \cmark & \cmark & \textbf{62.8} & \textbf{72.4} & \textbf{76.8} & \textbf{65.1} & 66.8 & \textbf{71.5} & \textbf{67.3} & 58.5 & \textbf{74.9} & \textbf{65.7} & \textbf{63.1} & \textbf{79.2} & \textbf{68.7} 
    \\
    \bottomrule
    \end{tabular}%
  }

  \label{tab:osda}%
  \vspace{-10pt}
\end{table*}%

\subsection{Domain Adaptation}
When DNNs encounter data from distributions that differ from their training data, a noticeable decline in performance is often observed. To tackle this issue, domain adaptation framework \cite{ganin2016domain, dayal2025leveraging, li2023source, kouw2019review, gao2021discrepancy} has been developed, which utilizes labeled data from source domains to train DNNs for unlabeled target domains using a transductive learning approach. We validate the effectiveness of \our{} using LEAD framework \cite{qu2024lead} on  OfficeHome \cite{venkateswara2017deep} dataset. For a fair comparison, we access source-only and augment data with \our{}. We experiment with different settings including across partial-set domain adaptation (PDA), open-set domain adaptation (OSDA), and open-partial-set domain adaptation (OPDA). The OPDA results are reported in Table~\ref{tab:opda_officehome}, while OSDA and PDA  performance evaluations are summarized in Tables~\ref{tab:osda} and \ref{tab:pda}, respectively.

\vspace{5pt}
\noindent{\textbf{Open-Partial Domain Adaptation (OPDA).}}  OPDA \cite{saito2021ovanet, you2019universal} falls under the broader scope of universal domain adaptation, where there is no prior knowledge about the label shift, such as common classes or the number of categories in the target domain. Some existing approaches \cite{saito2020universal, chen2022geometric} require simultaneous access to both source and target data which may become impractical due to data protection regulations \cite{voigt2017eu}. In Table \ref{tab:opda_officehome}, our baseline LEAD \cite{qu2024lead} has an average H-score of \numbersBlue{$75.0\%$}, while LEAD + \our{} achieves \numbersBlue{$77.2\%$}, representing a significant \numbersBlue{$2.2\%$} improvement in the average H-score. Particularly, for Ar2Pr, addition of our augmentation strategy improved performance of LEAD \cite{qu2024lead} by \numbersBlue{$4.9\%$}, for CL2Ar the improvement is \numbersBlue{$6.2\%$} and for Re2Ar improvement is \numbersBlue{$4.4\%$}. The results affirm that \our{} is an effective data augmentation for open partial domain adaptation scenarios. 

\par
\vspace{5pt}
\noindent{\textbf{Open-set Domain Adaptation (OSDA).}} In OSDA a significant portion of the target samples comes from new categories that are absent in the source domain. Since these categories do not have specific labels, existing methods often treat them collectively as a single "unknown" class.
In Table~\ref{tab:osda}, LEAD + \our{} achieved an average improvement in the H-score of \numbersBlue{$1.5\%$}  compared to the baseline LEAD, underscoring the significance of our proposed augmentation strategy.
Particularly,  considerable improvements of \numbersBlue{$4.1\%$} for Cl2Ar,  \numbersBlue{$5.4\%$} for Re2CL and \numbersBlue{$3.4\%$} for Re2Pr; can be observed by the addition of \our{} augmentation in the baseline LEAD \cite{qu2024lead} method.

\par
\vspace{5pt}
\noindent{\textbf{Partial Domain Adaptation (PDA).}} PDA focuses on adapting a model trained on a source domain to a target domain where only a subset of the classes overlap.
We compare the performance of a number of techniques on the Office-Home dataset while adapting LEAD as LEAD + \our to showcase the gain enabled by our method. 
Table~\ref{tab:pda} shows the accuracy for various domain adaptation tasks as well as the average accuracy across all tasks.  LEAD + \our{} achieved an average improvement in the H-score of \numbersBlue{$0.5\%$} over the  LEAD. Particularly, for Pr2Cl maximum improvement of \numbersBlue{$9.5\%$}, in Re2CL, an improvement of \numbersBlue{$4.6\%$}, in Cl2Pr, an improvement of \numbersBlue{$1.4\%$} is observed by the addition of \our{} augmentation.

\par
\begin{table*}[t]
  \centering
\caption{Accuracy comparison (\%) in the  Partial Domain Adaptation (PDA) scenario on the Office-Home dataset.}
\vspace{-5pt}
  \addtolength{\tabcolsep}{-2.0pt}
  \resizebox{0.99\textwidth}{!}{
    \begin{tabular}{lccc|ccccccccccccc}
    \toprule
\textbf{Method} & \textbf{U} & \textbf{SF} & \textbf{CF} & \textbf{Ar2Cl} & \textbf{Ar2Pr} & \textbf{Ar2Re} & \textbf{Cl2Ar} & \textbf{Cl2Pr} & \textbf{Cl2Re} & \textbf{Pr2Ar} & \textbf{Pr2Cl} & \textbf{Pr2Re} & \textbf{Re2Ar} & \textbf{Re2Cl} & \textbf{Re2Pr} & \textbf{Avg} \\

    \midrule
    CMU~\cite{fu2020_cmu} & \xmark  & \xmark & \cmark & 50.9  & 74.2  & 78.4  & 62.2  & 64.1  & 72.5  & 63.5  & 47.9  & 78.3  & 72.4  & 54.7  & 78.9  & 66.5 \\
    DANCE~\cite{saito2020_unida_dance} & \cmark & \xmark & \cmark & 53.6  & 73.2  & 84.9  & 70.8  & 67.3  & 82.6  & 70.0  & 50.9  & 84.8  & 77.0  & 55.9  & 81.8  & 71.1 \\
    DCC~\cite{li2021_dcc}  & \cmark & \xmark & \xmark & 54.2  & 47.5  & 57.5  & 83.8  & 71.6  & 86.2  & 63.7  & 65.0  & 75.2  & 85.5  & 78.2  & 82.6  & 70.9 \\
    OVANet~\cite{saito2021_ovanet} & \xmark & \xmark & \cmark & 34.1  & 54.6  & 72.1  & 42.4  & 47.3  & 55.9  & 38.2  & 26.2  & 61.7  & 56.7  & 35.8  & 68.9  & 49.5 \\
    GATE~\cite{chen2022_gate} & \cmark & \xmark & \cmark & 55.8  & 75.9  & 85.3  & 73.6  & 70.2  & 83.0  & 72.1  & 59.5  & 84.7  & 79.6  & 63.9  & 83.8  & 74.0 \\
    Source-only &- &- &- & 45.9  & 69.2  & 81.1  & 55.7  & 61.2  & 64.8  & 60.7  & 41.1  & 75.8  & 70.5  & 49.9  & 78.4  & 62.9 \\
    \midrule
    LEAD \cite{qu2024lead} & \cmark & \cmark & \cmark & 58.2 & 83.1 & 87.0 & 70.5 & 75.4 & 83.3 & 73.7 & 50.4 & 83.7 & 78.3 & 58.7 & 83.2 & \textbf{73.8} 
    \\
    + \textbf{\our{}}  & \cmark & \cmark & \cmark & \textbf{58.9} & 76.3 & \textbf{87.7} & \textbf{71.3} & \textbf{76.8} & \textbf{83.4} & \textbf{73.9} & \textbf{59.2} & \textbf{81.9} & \textbf{79.1} & \textbf{63.3} & 80.6 & \textbf{74.3}  \\

    \bottomrule
    \end{tabular}%
  }
  \label{tab:pda}%
\end{table*}


\begin{table}
\centering
\caption{Top-1 (\%) accuracy of self-supervised learning methods. Adding \our{}{} yields better performance.}
\vspace{-5pt}
\begin{tabular}{lccc}
\toprule
\textbf{ Method} & \textbf{ Flower102} & \textbf{ Stanford Cars} & \textbf{ Aircraft} \\
\midrule
MoCo v2  \cite{he2020momentum} & 80.31 &  40.82 &  51.36 \cr
DiffuseMix \cite{islam2024diffusemix} & 82.15   & 41.73 & 53.28 \cr
\textbf{\our{}}  & \textbf{83.57} & \textbf{43.71}  & \textbf{54.68}  \\
\midrule
SimSiam  \cite{chen2021exploring} & 86.93  & 48.34 & 40.37 \cr

DiffuseMix \cite{luo2023camdiff} &  89.24  & 49.17 & 42.63 \cr
\textbf{\our{}}  & \textbf{90.45} & \textbf{ 51.78}  & \textbf{43.75} \\ 
\bottomrule
\end{tabular}
\label{tab:ssl_experiments}
\end{table}

\subsection{Self-Supervised Learning}
We experimented with self-supervised learning models MoCo v2 \cite{he2020momentum} and SimSiam \cite{chen2021exploring} by incorporating our proposed  \our{} augmentation method. 
Table \ref{tab:ssl_experiments} showcases the Top-1 accuracy of self-supervised learning methods, comparing the performance of MoCo v2 and SimSiam, both with and without augmentation. On the Flower102, Stanford Cars, and Aircraft datasets, the baseline MoCo v2 achieves accuracies of \our{} delivers absolute gains of \numbersBlue{$1.42\%$}, \numbersBlue{$1.98\%$} and \numbersBlue{$1.40\%$} on these datasets compared to DiffuseMix \cite{islam2024diffusemix}. Similarly, SimSiam baseline achieved accuracies of \numbersBlue{$86.93\%$}, \numbersBlue{$48.34\%$}, and \numbersBlue{$40.37\%$} on the Flower102, Stanford Cars, and Aircraft datasets. Compared with DiffuseMix \cite{islam2024diffusemix}, augmenting it with the proposed method gives absolute gains of \numbersBlue{$1.21\%$},  \numbersBlue{$2.61\%$} and \numbersBlue{$1.12\%$}. These results  illustrate the efficacy of \our{}, as it consistently boosts the performance across various datasets for self-supervised learning methods. The systematic gains highlight the robustness of our approach, demonstrating its potential to enhance the accuracy of state-of-the-art models in different domains.

\section{Analysis and Discussion}
In this section, we present  an in-depth analysis of the different design choices made in \our{}, supporting it by a  comprehensive ablation study.





\subsection{Grad-CAM Visualization}
\begin{figure*}[t]
    \centering
    \includegraphics[width=0.90\linewidth]{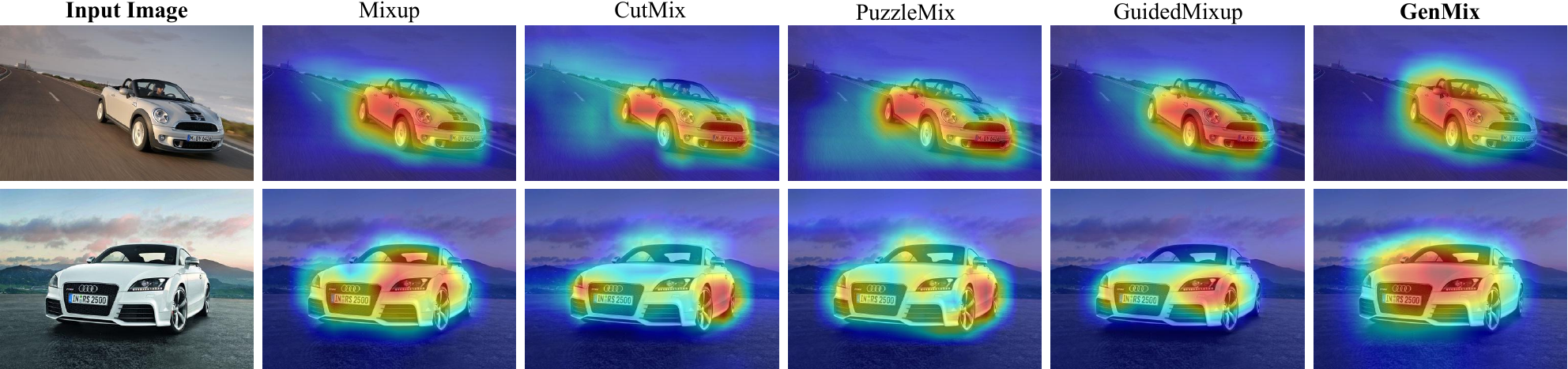}
    \caption{Representative Grad-CAM visualizations of saliency maps for Mixup \cite{mixup}, CutMix \cite{cutmix}, PuzzleMix \cite{kim2020puzzle}, GuidedMixup \cite{kang2023guidedmixup} methods on random input images of Stanford Cars dataset. We train ResNet50 from scratch for a fair comparison. Saliency maps for \our{} are more consistent and better focused on the salient image regions.}
    \vspace{-10pt}
    \label{fig:gradcam}
\end{figure*}

Gradient-weighted Class Activation Mapping (Grad-CAM) \cite{selvaraju2017grad} is used to visualize class-discriminative regions in input images. It works by computing the gradients of the target class with respect to the feature maps in the last convolutional layer, highlighting the important regions contributing to the prediction. This helps in understanding where the model is focusing to make decisions, making it useful for interpretability.
Using models trained from scratch, comparison with four existing methods \cite{mixup, cutmix, kim2020puzzle, kang2023guidedmixup, uddin2020saliencymix} using Grad-CAM visualization  is presented in Figure \ref{fig:gradcam}. We observe that the models with existing mixup based methods  have less focused region in some images, compared to our proposed \our. It shows that these methods focus specific object features such as front lights or wheels of the car only.  The Grad-CAM saliency map of the model trained using \our{}  focuses on the full object, implying better generalization. 
This property can be  attributed to the fact that \our{} works with a single image and mixes different parts of the same image and its edited versions to augment the data. Our method covers more area of the object while  focusing less on the context, whereas the compared approaches are more local within the object and in some cases also have  focus outside the object. 
Thus \our{} better retains the structure of the foreground object, resulting in an enlarged focus  on the entire object representing better generalization.


\vspace{5pt}
\subsection{Ablation Studies} 
To assess the significance of the component of our approach, we performed an ablation study by systematically removing the impact of individual components and recording the performance on ResNet-50 across the Stanford Cars and Flowers-102 datasets, as shown in Table \ref{tab:ablation}.
\par
In the table, the baseline \cite{resnet} using only the original images $I_i$ obtains Top-1 and Top-5 accuracies of \numbersBlue{$85.52\%$} on Stanford Cars and \numbersBlue{$78.83\%$} on Flowers102. Introducing fractal interpolation ($F_v$) to the input images brings a slight improvement in performance for both datasets. We then eliminate both seamless concatenation ($H_{iju}$) and fractal interpolation ($F_v$) by conducting experiments using only the edited images ($\hat{I}_{ij}$) as augmented images for training. This setup aligns more closely with methods in \cite{trabucco2023effective,azizi2023synthetic}, which also use diffusion-generated images as augmentations. Our results for this experiment in the table are consistent with the insights of~\cite{azizi2023synthetic}, suggesting that using generated images directly does not always significantly outperform the vanilla approach. We then perform experiments where edited images are interpolated with fractal images during training. In the Cars dataset, this leads to a slight improvement over the baseline, with Top-1  accuracies of \numbersBlue{$89.72\%$}. However, the Flowers dataset shows a slight drop in performance, indicating that fractal interpolation is more effective when combined with greater diversity in the training set. Next, we remove the fractal interpolation but retain the seamless concatenation of edited ($\hat{I}_{ij}$) and original ($I_i$) images to form hybrid images ($H_{iju}$) for data augmentation. This boosts performance for both datasets. We conjecture that this improvement is due to the combined presence of both generated and original image content in each augmented image, highlighting the importance of the seamless concatenation step in \our{}. Finally, when all components,
including fractal interpolation, are incorporated, the model achieves its best performance for both datasets. These consistent improvements with each additional component emphasize the effectiveness of the design choices in \our{} for data augmentation.
\begin{table}
\centering
\caption{Ablation study using Stanford Cars (Cars) and Flowers102 (Flowers) datasets. Top-1 and Top-5 accuracies are reported with different combinations of \(I_{i}\): Input image, \(\hat{I}_{ij}\): Edited images using prompts \(p_j\), \(H_{iju}\): Hybrid images using random mask \(M_u\), and \(F_{v}\): fractal images used to obtain final interpolated image \(A_{ijuv}\).}
\begin{tabular}{cl|ccccccc}
& \(I_i\) & \ding{51} & \ding{51}   & -  & - & -  & -  \\
& \(\hat{I}_{ij}\)  & - & -   & \ding{51}  &\ding{51}  & -  & - \\
& \(H_{iju}\)  & - &- & -  & -  & \ding{51} & \ding{51}  \\
& \(F_{v}\) & - &\ding{51} & - &\ding{51} & - & \ding{51} \\ \midrule
\multirow{1}{*}{\rotatebox{0}{Cars}} 
& Top-1  & 85.52 & 86.73   & 88.13 & 89.72 & 91.87 & \textbf{92.48}  \\
\midrule
\multirow{1}{*}{\rotatebox{0}{Flowers}} 
& Top-1  & 78.73 & 78.34   & 77.88 & 77.81 & 90.67 & \textbf{81.56}  \\
\bottomrule
\end{tabular}
\label{tab:ablation}
\end{table}

\par
\vspace{5pt}
\subsection{Augmentation Overhead} 
\label{sec:augmentation_overhead_supp} 
\begin{figure}
    \centering
    \includegraphics[width=1.0\linewidth]{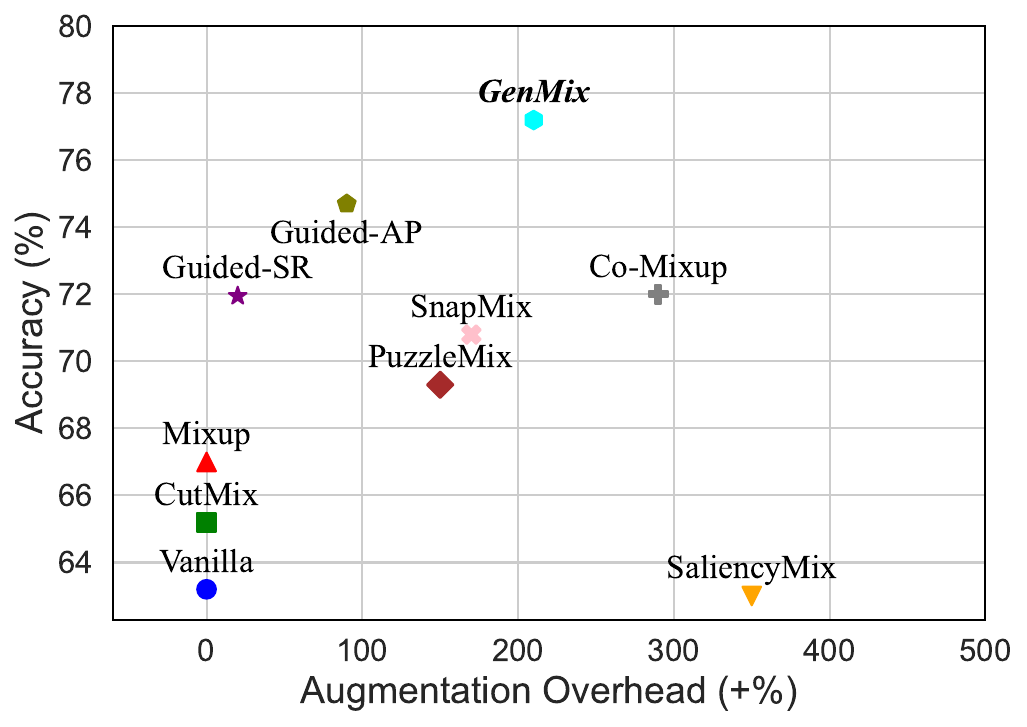}
    \caption{Augmentation overhead $\mathcal{A_{O}}$ vs Accuracy (\%) plot on CUB-200-2011 dataset with batch size 32.} 
    \vspace{-10pt}
    \label{fig:augOverhead}
\end{figure}

We evaluate the computational overhead of various state-of-the-art image augmentation methods, including \our{}{}, in relation to their performance improvements. As defined by Kang \etal \cite{kang2023guidedmixup}, the augmentation overhead $\mathcal{A_{O}}$ is calculated as:
$$\mathcal{A_{O}} = \frac{\mathcal{T}_{aug} - \mathcal{T}_{van}}{\mathcal{T}_{van}} \times 100 (\%),$$
where $\mathcal{T}_{aug}$ refers to the total time required for image generation and training, and $\mathcal{T}_{van}$ represents the training time of the baseline model without augmentation. Although \our{} could leverage parallel  processing to speed up image generation, in this experiment, we used sequential processing to ensure a fair comparison.  As apparent in Fig.~\ref{fig:augOverhead},  \our{} achieves an excellent balance between performance and augmentation overhead, surpassing all other methods in accuracy while maintaining significantly lower overhead compared to Co-Mixup \cite{kim2020co} and SaliencyMix \cite{uddin2020saliencymix}. Additionally, \our{} can be further optimized by saving the generated images offline, enabling quicker execution of multiple training experiments, which is particularly useful for research and optimization purposes.

\par

\subsection{Impact of Prompt Selection}
\label{supp:bad_prompts} 
\begin{figure}
    \centering
    \includegraphics[width=0.95\linewidth]{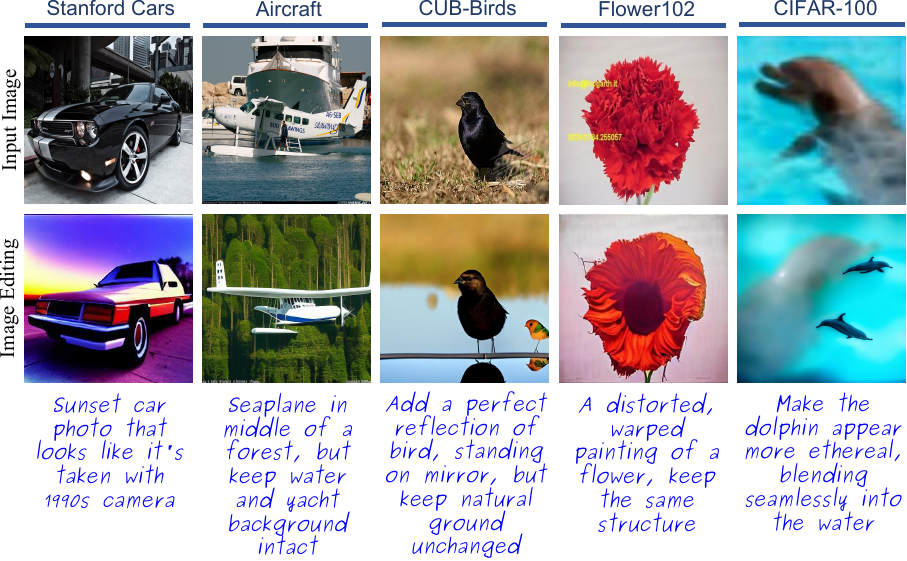}
\caption{ Some examples of badly designed prompts: \textbf{Top row:} Original training examples.  \textbf{Bottom row:} The corresponding generated images. It demonstrates that using detailed prompts (in blue) leads to poor quality images that are unsuitable for training. The image in the last column from CIFAR100, along with its prompt, produces a \textbf{\emph{{black}}} image with no visible output.} 
\label{fig:poor_prompts}
\end{figure}

Diffusion model outputs are  dependent on the input prompts \cite{du2023stable}. Our approach to designing \emph{bespoke conditional prompts} focuses on creating prompts that edit images while maintaining their structural integrity, making them applicable across various datasets. As outlined in the main text, we introduce \textit{filter-like} prompts such as snowy, sunset, and rainbow, demonstrating their effectiveness in training robust classifiers.
We also explore some \textit{bad} prompts that are unsuitable for the image editing  in \our{} as shown in Figure \ref{fig:poor_prompts}. Descriptive or overly complex prompts tend to produce images that deviate significantly from the original distribution, often leading to unrealistic foregrounds and backgrounds, making them ineffective for training. It further emphasizes the value of our proposed \textit{filter-like bespoke conditional prompts}, which avoid introducing unwanted alterations to the training data.



\par
\vspace{5pt}
\subsection{Increasing Smooth Mask Vs Performance} Experiments were also conducted using the Flowers102 dataset to assess how different smooth masking strategies affect the overall performance of the proposed method. The results are reported in Table~\ref{tab:mask_ablation}. Employing even a single type of mask - specifically, vertical mask; our approach significantly outperforms the vanilla \cite{resnet} baseline in terms of accuracy. Introducing both vertical and horizontal masks leads to further improvements. The highest accuracy is achieved when both mask types are combined with random flipping between the positions of input and generated images, enhancing the diversity of the training data. Moreover, the method is compatible with masking techniques from previous studies, such as those in \cite{cutmix, mixup}.

\begin{table}

\centering
\caption{Effects of masking  in \our{}~on  Flower102. All variants yield notably superior results compared to the vanilla. However, best results are achieved when horizontal (Hor) and vertical (Ver) and patchswap masks are used with vice versa.} 
\begin{tabular}{lcc}
\toprule
\multicolumn{1}{l}{ \textbf{Mask} } & \multicolumn{1}{c}{ \textbf{Top-1 (\%)}} \\
\midrule
Vanilla~\cite{resnet} & 89.74  \\ 
Ver Mask ( \begin{tikzpicture}[baseline=-0.125ex]
            \shade[top color=black, middle color=white, bottom color=white, shading angle=90] (0,0) rectangle (0.25cm,0.25cm);
            \draw (0,0) rectangle (0.25cm,0.25cm); 
          \end{tikzpicture}
) & 94.11 \\
Hor + Ver Masks ( 
\begin{tikzpicture}[baseline=-0.125ex]
            \shade[bottom color=black, middle color=black!50, top color=white, shading angle=90] (0,0) rectangle (0.25cm,0.25cm);
            \draw (0,0) rectangle (0.25cm,0.25cm); 
          \end{tikzpicture},
 \begin{tikzpicture}[baseline=-0.125ex]
            \shade[left color=black, middle color=black!50, right color=white, shading angle=0] (0,0) rectangle (0.25cm,0.25cm);
            \draw (0,0) rectangle (0.25cm,0.25cm); 
          \end{tikzpicture}
) & 94.23  \\
Hor + Ver + Flipping ( \begin{tikzpicture}[baseline=-0.125ex]
            \shade[bottom color=black, middle color=black!50, top color=white, shading angle=90] (0,0) rectangle (0.25cm,0.25cm);
            \draw (0,0) rectangle (0.25cm,0.25cm); 
          \end{tikzpicture},
           \begin{tikzpicture}[baseline=-0.125ex]
    \shade[left color=black, middle color=black!50, right color=white, shading angle=0] (0,0) rectangle (0.25cm,0.25cm);
    \draw (0,0) rectangle (0.25cm,0.25cm); 
\end{tikzpicture},
          
          \begin{tikzpicture}[baseline=-0.125ex]
            \shade[left color=black, middle color=black!50, right color=white, shading angle=0] (0,0) rectangle (0.25cm,0.25cm);
            \draw (0,0) rectangle (0.25cm,0.25cm); 
          \end{tikzpicture},
          
          \begin{tikzpicture}[baseline=-0.125ex]
            \shade[bottom color=black, middle color=black!50, top color=white, shading angle=90] (0,0) rectangle (0.25cm,0.25cm);
            \draw (0,0) rectangle (0.25cm,0.25cm); 
            \end{tikzpicture})  
            & 95.55  
\\

Hor + Ver + Flipping  + PatchSwap ( 
\begin{tikzpicture}[baseline=0.05cm]
  \fill[black] (0,0) rectangle (0.25cm, 0.25cm); 
  \fill[white, rounded corners=0.02cm, opacity=0.8]
    (0.05cm,0.05cm) rectangle (0.20cm,0.20cm);
  \draw (0,0) rectangle (0.25cm, 0.25cm); 
\end{tikzpicture},

\begin{tikzpicture}[baseline=0.05cm]
  \fill[white] (0,0) rectangle (0.25cm, 0.25cm); 
  \fill[black, rounded corners=0.02cm, opacity=0.8]
    (0.05cm,0.05cm) rectangle (0.20cm,0.20cm);
  \draw (0,0) rectangle (0.25cm, 0.25cm); 
\end{tikzpicture}) & 96.26  

\\
\bottomrule
\end{tabular}
\label{tab:mask_ablation}
 \vspace{-10pt}
\end{table}

\section{Conclusion}
\label{sec:conclusion}
In this work,  a data augmentation technique dubbed as \our{} is proposed which is built on pre-trained image editing diffusion models. \our{} enhances diversity in the training data while preserving the original semantic content of the input images. It employs \emph{editing, smooth concatenation, and fractal blending} to produce the final augmented images. \our{} has shown performance improvement across multiple tasks including general classification, fine-grained classification, domain adaptation, addressing data scarcity, fine-tuning, self-supervised learning, and adversarial robustness, over several architectures and benchmark datasets including 
\emph{ImageNet-1k, Tiny-ImageNet-200, CIFAR-100, Oxford Flower102, Caltech Birds, Stanford-Cars, FGVC Aircraft, and Office-Home}. Across a wide range of experiments, our method has consistently demonstrated improved performance outperforming existing SOTA image augmentation methods. The proposed method has the potential to improve learning of deep models across a wide range of tasks. 

\vspace{0.7mm}
\noindent\textbf{Limitations:} 
\our{} has two key limitations: (1) It relies on text prompts for image editing, and inappropriate text prompts may result in undesired outputs. To address this, we propose a set of \textit{filter-like} prompts that can be applied to a wide range of natural images. 
(2) \our{} introduces additional overhead for editing images, but this cost may be considered as a trade off for improved convergence in large-scale learning models. Image editing cost can be effectively handled by generating edited images in advance and employing faster image editing methods.

\small
\bibliographystyle{IEEEtran}
\bibliography{main}
\begin{IEEEbiography}[{\includegraphics[width=1in,height=1.25in,clip,keepaspectratio]{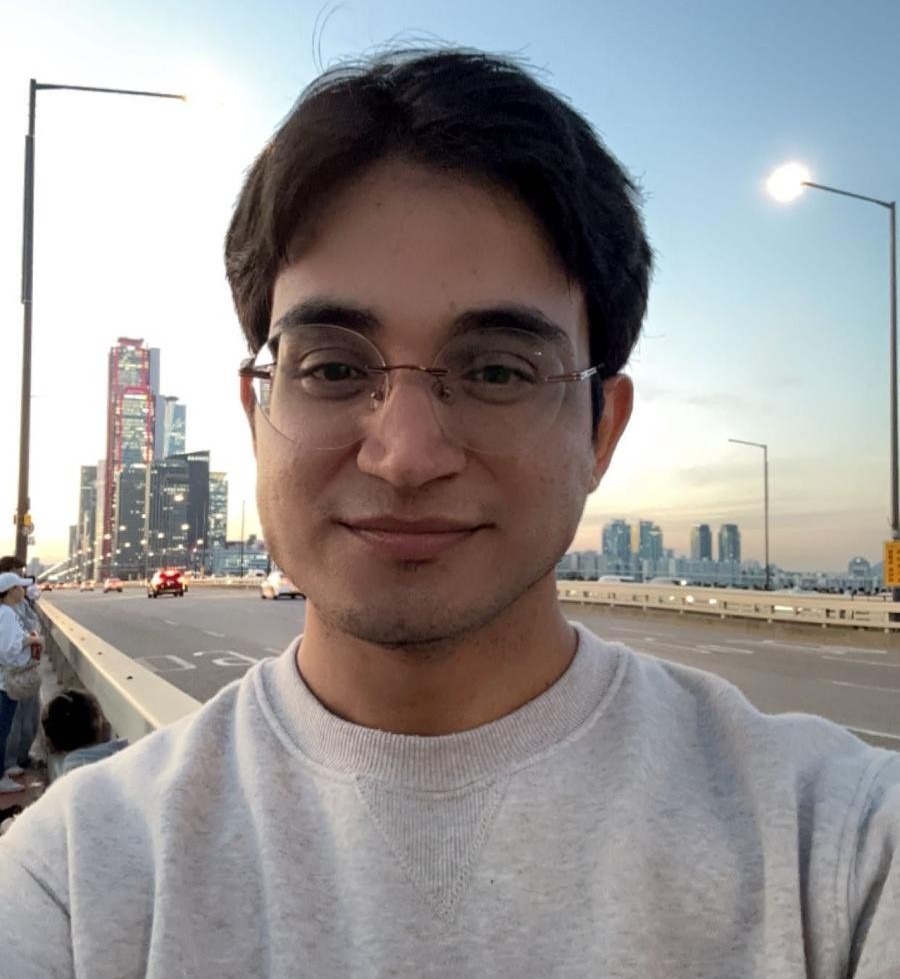}}]{Khawar Islam} is a Senior Research Scientist specializing in Generative AI, currently with Upendi.app, Seoul, South Korea. Currently, I am exploring various techniques such as continual learning, mixup, style transfer, and data augmentation for enhancing model robustness. I am working on numerous Generative AI techniques such as Diffusion, CosXL, and FLUX. I have also been involved in large-scale projects, including dataset construction, large language models, vision-language models in my work. With an academic background in computer engineering, I have earned my MS degree from Sejong University, I have contributed to various research domains including sewer defect classification, image compression, face recognition, and vision transformers.
\end{IEEEbiography}

\begin{IEEEbiography}[{\includegraphics[width=1in,height=1.25in,clip,keepaspectratio]{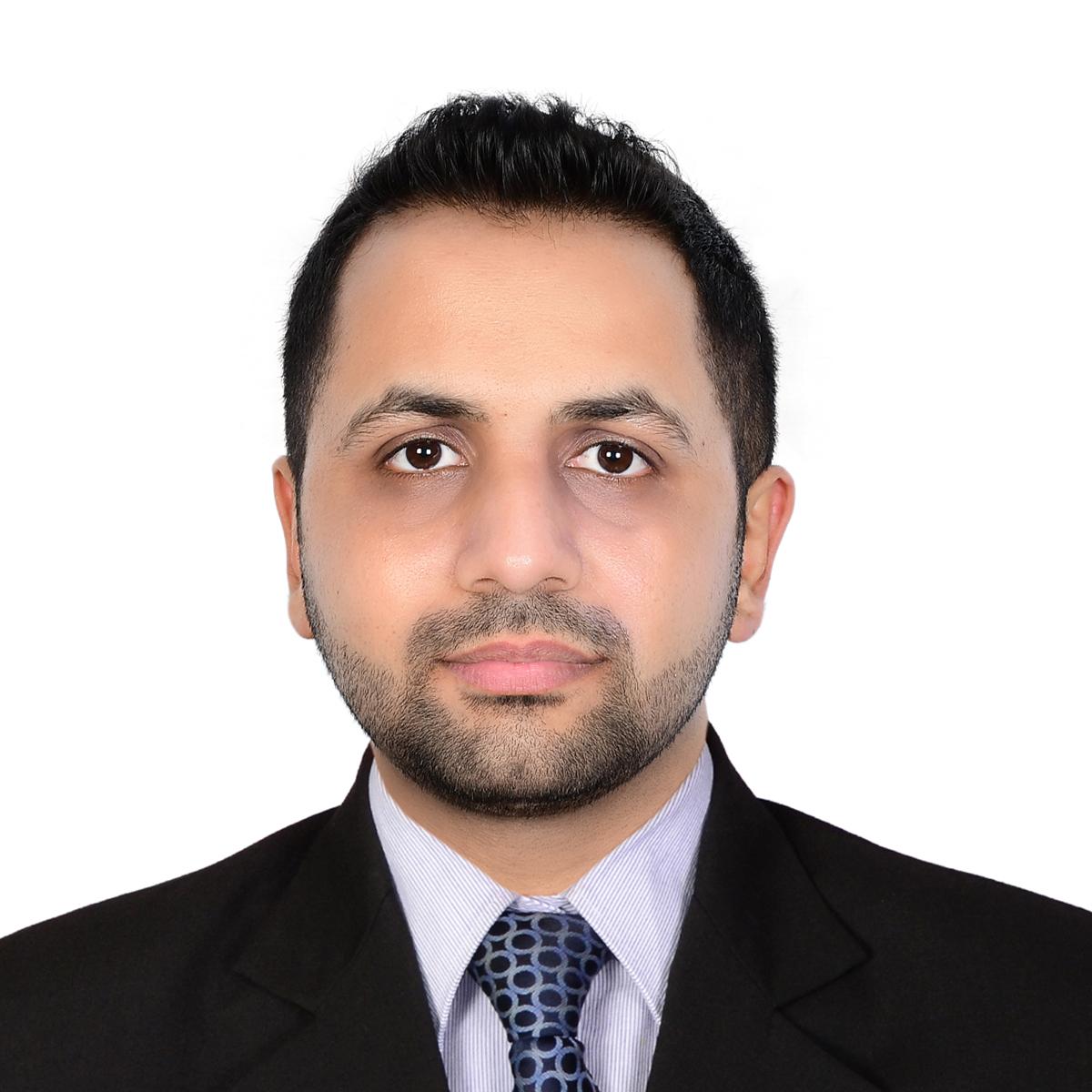}}]{Muhammad Zaigham Zaheer}
is currently associated with Mohamed bin Zayed University of Artificial Intelligence as a Research Fellow. Previously, he has worked with the Electronics and Telecommunications Research Institute (ETRI), Ulsan, Korea, as a post-doctoral researcher.
He received his PhD from the Korea University of Science and Technology (UST) in 2022 and his MS degree from Chonnam National University in 2017. His research interests include computer vision,  anomaly detection in images/videos, and semi-supervised/self-supervised/unsupervised learning.
\end{IEEEbiography}
\begin{IEEEbiography}[{\includegraphics[width=1in,height=1.25in,clip,keepaspectratio]{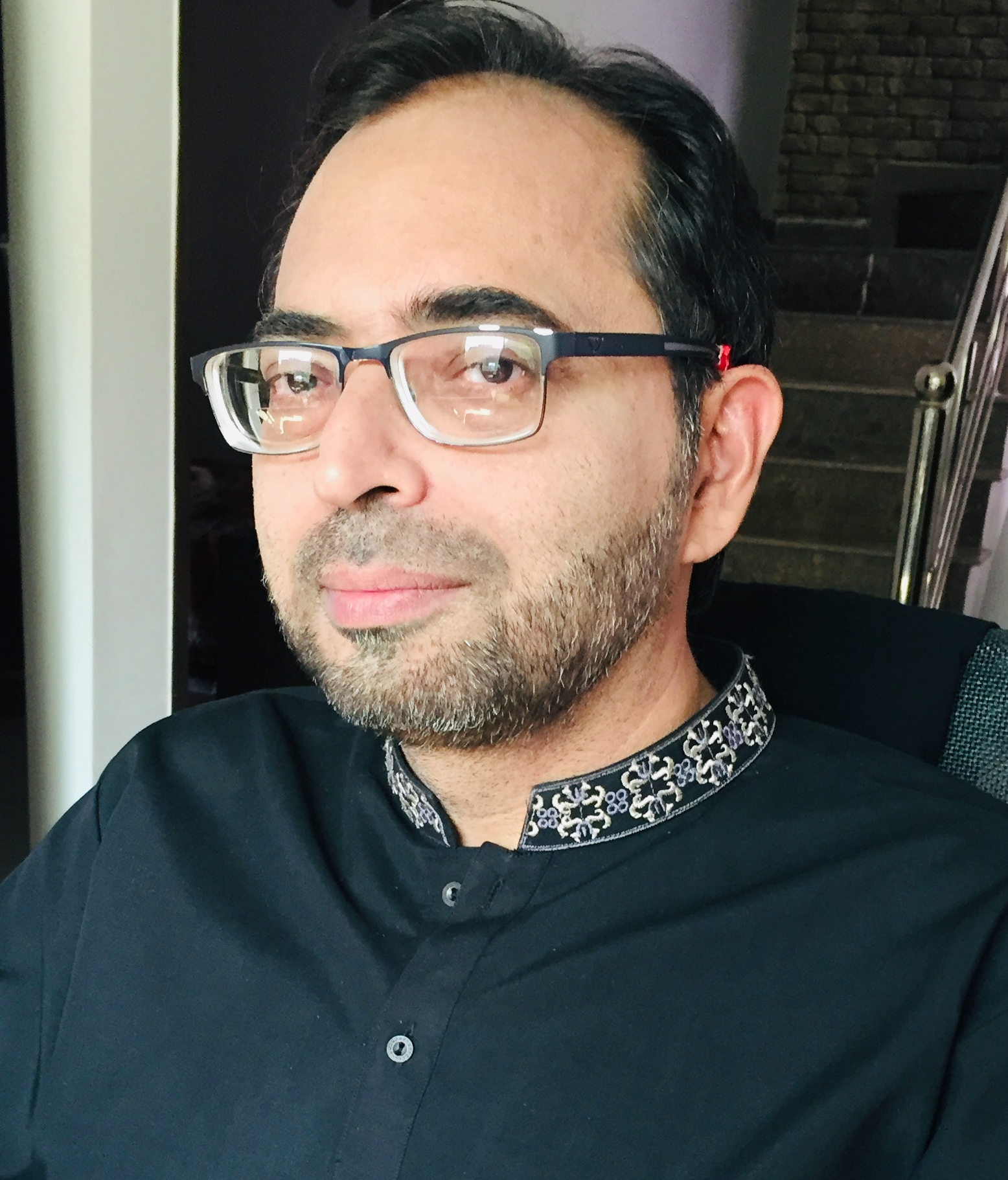}}]{Arif Mahmood} is a Professor of Computer Science and Dean Faculty of Sciences at Information Technology University of Punjab (ITU). He is also the Director of the Center for Robot Vision at ITU. His current research directions in Computer Vision are person pose detection and segmentation, crowd counting and flow detection, background-foreground modeling in complex scenes, object detection, human-object interaction detection, and abnormal events detection. He is also actively working in diverse Machine Learning applications including  cancer grading and prognostication using histology images, and environmental monitoring using remote sensing. He has also worked as a Research Assistant Professor with the School of Mathematics and Statistics, University of Western Australia (UWA) where he worked on community detection in complex networks. Before that, he was a Research Assistant Professor at the School of Computer Science and Software Engineering and performed research on human face recognition, object classification, and action recognition. 
\end{IEEEbiography}
\begin{IEEEbiography}[{\includegraphics[width=1in,height=1.25in,clip,keepaspectratio]{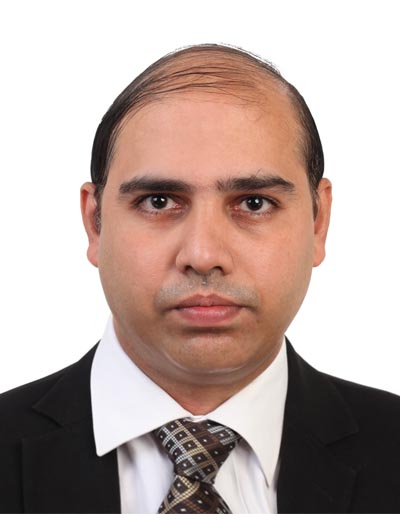}}]{Karthik Nandakumar} is an Associate Professor in the Computer Vision department at Mohamed bin Zayed University of Artificial Intelligence (MBZUAI). Prior to joining MBZUAI, he was a Research Staff Member at IBM Research – Singapore and a Scientist at Institute for Infocomm Research, A*STAR, Singapore. His research interests include computer vision, machine learning, biometric recognition, and applied cryptography. Specifically, he is interested in developing machine learning algorithms for biometrics and video surveillance applications as well as various security and privacy-related issues in machine learning. He is a Senior Area Editor of IEEE Transactions on Information Forensics and Security (T-IFS) and was a Distinguished Industry Speaker for the IEEE Signal Processing Society (2020-21). 
\end{IEEEbiography}
\begin{IEEEbiography}[{\includegraphics[width=1in,height=1.25in,clip,keepaspectratio]{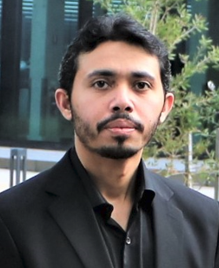}}]{Naveed Akhtar} is a Senior Lecturer  at the University of Melbourne. He received his PhD in Computer Science from the University of Western Australia and Master degree from Hochschule Bonn-Rhein-Sieg, Germany. He is a recipient of the Discovery Early Career Researcher Award from the Australian Research Council. He is a Universal Scientific Education and Research Network Laureate in Formal Sciences, and a recipient of Google Research Scholar Program award in 2023. He was a finalist of the Western Australia's Early Career Scientist of the Year 2021.  He is an ACM Distinguished Speaker and serves as an Associate Editor of IEEE Trans. Neural Networks and Learning Systems. He has served as an Area Chair for reputed conferences like IEEE Conf. on Computer Vision and Pattern Recognition (CVPR) and European Conference on Computer Vision (ECCV) on multiple occassions.
\end{IEEEbiography}

\end{document}